\documentclass[times,twocolumn,final,authoryear]{elsarticle}
\usepackage{graphicx}
\usepackage{filecontents,lipsum}

\usepackage{comment}
\usepackage{amssymb}
\usepackage{amsmath}
\usepackage{mathtools}
\usepackage{breqn}
\usepackage{varwidth}
\usepackage[most]{tcolorbox}
\usepackage{array}
\usepackage{multirow}
\usepackage{footnote}
\usepackage{cases}
\usepackage{listings}
\usepackage{booktabs}
\usepackage{tabularx}
\usepackage{breqn}
\usepackage{float}
\usepackage{tikz}
\usepackage{tabularx}
\usepackage[flushleft]{threeparttable}

\usepackage{afterpage}
\usepackage{tikz}
\usepackage{lipsum}
\usepackage{capt-of}
\usepackage{tikz}
\usepackage{scrextend}
\usepackage{mathrsfs}
\usetikzlibrary{arrows}
\usepackage{epsfig}	
\usepackage{booktabs}
\usepackage[belowskip=-1pt,aboveskip=0pt]{caption}
\usepackage{subcaption}
\usepackage{breqn}
\usepackage{array}

\newcolumntype{P}[1]{>{\centering\arraybackslash}p{#1}}
\newcolumntype{M}[1]{>{\centering\arraybackslash}m{#1}}

\DeclareGraphicsExtensions{.pdf,.jpeg,.png,.tif}
\makeatletter
\def\BState{\State\hskip-\ALG@thistlm}
\usepackage[figuresright]{rotating}	

\usepackage{xcolor}
\usepackage{color, colortbl}
\usepackage[linesnumbered,ruled,vlined]{algorithm2e}

\SetCommentSty{mycommfont}
\usepackage{algcompatible}
\usepackage[ruled,vlined]{algorithm2e}
\usepackage{algpseudocode}

\newcommand{\mycc}{\cellcolor{lightgray}}
\newcommand{\myrr}{\cellcolor{red!15}}
\newcommand{\mybb}{\cellcolor{blue!15}}

\definecolor{Gray}{gray}{0.4}
\usepackage[pagebackref=false,breaklinks=true,letterpaper=true,colorlinks,bookmarks=true]{hyperref}

\usepackage[utf8]{inputenc}
\usepackage[english]{babel}
\usepackage{prletters}
\usepackage{framed,multirow}

\usepackage{amssymb}
\usepackage{latexsym}
\usepackage[utf8]{inputenc}
\usepackage[english]{babel}
\usepackage{url}
\usepackage{xcolor}
\definecolor{newcolor}{rgb}{.8,.349,.1}

\newcommand\T{\rule{0pt}{5.6ex}}       

\journal{Pattern Recognition Letters}
\hypersetup{draft}
\begin{document}

	\ifpreprint
	\setcounter{page}{1}
	\else
	\setcounter{page}{1}
	\fi
	
	\begin{frontmatter}
		
		
		\title{Exploring Multi-Tasking Learning in Document Attribute Classification}
		
		\author[1]{Tanmoy \snm{Mondal}\corref{cor1}} 
		\ead{tanmoy.besu@gmail.com}
		\author[2]{Abhijit \snm{Das}\corref{cor2}}
		\ead{abhijitdas2048@gmail.com }
		\author[3]{Zuheng \snm{Ming}}
		
		\address[1]{SC Team, IMT Atlantique, Brest, France}
		\address[2]{Thapar University, Punjab, India}
		\address[3]{L3i, University of La-Rochelle, France}
		

		\begin{abstract}
			In this work, we adhere to explore a Multi-Tasking learning (MTL) based network to perform document attribute classification such as the font type, font size, font emphasis  and  scanning  resolution classification of a document image. To accomplish these tasks, we operate on either segmented word level or on uniformed size patches randomly cropped out of the document. Furthermore, a hybrid convolution neural network (CNN) architecture  "MTL+MI", which is based on the combination of MTL and Multi-Instance (MI) of patch and word is used to accomplish joint learning for the classification of the same document attributes. The contribution of this paper are three fold: firstly, based on segmented word images and patches, we present a MTL based network for the classification of a full document image. Secondly, we propose a MTL and MI (using segmented words and patches) based combined CNN architecture (``MTL+MI") for the classification of same document attributes. Thirdly, based on the multi-tasking classifications of the words and/or patches, we propose an intelligent voting system which is based on the posterior probabilities of each words and/or patches to perform the classification of document's attributes of complete document image.
		\end{abstract}
		
		
	\end{frontmatter}
	
	
	\section{Introduction}
	\label{sec1}
	
	Automatic analysis of document attributes such as font is highly useful for several document processing tasks such as character recognition \citep{Simard2003}, document classification, writer and script identification \citep{Shi2016}. 
	Automatic identification of font type, font size and font emphasis can highly improve the accuracy of \emph{Optical Character Recognition (OCR)} systems, especially when the data is processed in multi-script or multi-language form. 
	If text lines or words are labeled with a font class, then a specialist OCR system can potentially achieve better recognition rates than an OCR system trained on many fonts.
	
	Contrary to previous techniques in the literature, which have been mainly designed for a single task such as \emph{font type} classification \citep{Tensmeyer2017}, \citep{Cloppet2018}, this work focuses to learn multiple attributes in a multi-tasking environment instead of using a single network for each attribute classification. 
	Given a document image, the motivation of this work is to automatically identify the Font type (e.g \emph{Arial}, \emph{Calibri}, \emph{Courier}, \emph{Times new roman}, \emph{Trebuchet}, \emph{Verdana}), Font size (e.g. $8$, $10$, $12$), Font emphasis (e.g. \emph{bold}, \emph{italic}, \emph{bold-italic}, \emph{none}) and scanning resolution of the document (e.g. $150$ dpi, $300$ dpi, $600$ dpi) image by using a single \emph{MTL-based convolutional neural network (CNN)}.
	
	To handle the variations of various font characteristics, we compute deep features from the segmented word images and also from the extracted patches. In our fist approach, we have applied a MTL system for the classification of document attributes by using either word image or patch images as an input. In our second approach, we have adapted a hybrid CNN of two stream to perform the same multi-tasks. In the first stream, features are extracted from the input images from pre-trained CNN \citep{Tensmeyer2017} and in second stream, sequential patches are cut from the complete document image and the features are extracted from these patches 
	These two streams act as multi-instances, which are combined by vector-wise operation. We term this proposed MTL and MI based architecture as MTL+MI.
	The differences among different font types are subtle or even tiny and to capture these differences, we need to operate on local level and that's why we adopted to work on segmented words and/or patches. In addition, we proposed a voting method for the classification of each word and/or patch, which are used as the candidates for voting to decide the final class of the whole document image. 
	
	In summary, the contribution of the paper are as follows: i) we present a MTL based framework to classify the font type, font size, font emphasis and scanning resolution of document images. ii) a MTL+MI based framework to jointly learn the tasks with multi-instances i.e. using segmented word and cropped patch images together, iii) we have proposed an intelligent voting system based on the posterior probabilities to perform the classification of the complete document image. The code and dataset used in this research work, can be found in {\color{blue}\url{https://github.com/tanmayGIT/Document-Attribute-Classification.git}} and in {\color{blue}\url{http://navidomass.univ-lr.fr/TextCopies/}} respectively.

	\section{Related Works}
	\label{related_work}
	We review the literature in two direction: the first one is Font type/family and Font size classification and the second one is on multi-tasking network, applied in different domains of computer vision.
	The existing techniques in the literature about font type/family recognition can be divided into two main family. One is \emph{holistic approach} and other one is \emph{training based approaches}, mainly using deep neural network such as CNN and Recurrent Neural Network (RNN) based techniques. 
	A method proposed by \citep{BenMoussa2010} for Arabic font recognition by using \emph{Fractal geometry}, which has resulted in $98\%$ accuracy for $10$ font classes. 
	Another technique is proposed for optical font recognition by using typographical features and by using multivariate Bayesian classifier in \citep{BenMoussa2010}. This approach reported an accuracy of $97.35\%$ over English text lines for $10$ font classes.    
	More recently, deep learning techniques based on CNN and RNN have shown very high potential for font classification. 
	By considering font recognition on a single Chinese character is a sequence of classification problem, the authors in \citep{Tao2016} has proposed 
	principal component based 2D long short-term memory (LSTM) algorithm and were able to classify single Chinese characters into $7$ font classes with $97.77\%$ accuracy. 
	Identification of scripts in natural images was proposed in \citep{Shi2016}. The basic idea is combining deep features and mid-level representations into a globally trainable deep model. 
	The classification of hand-written Chinese characters into $5$ calligraphy classes is performed in \citep{Pengcheng2017}. They have obtained $95\%$ accuracy by using deep features, extracted from a pre-trained CNN on natural images. 
	A competition on the classification of medieval handwriting in Latin script was organized in \citep{Cloppet2018}. 
	The top performing technique obtained an accuracy of $83.9\%$ among $7$ methods, submitted in the competition. 
	One recent approach by \citep{Tensmeyer2017} presents a simple framework based on CNNs. 
	Their method achieved a state-of-the art performance on challenging data-set of $40$ Arabic computer fonts with $98.8\%$ line level accuracy.
	
	
	Multi-task Learning (MTL) exploits the task relatedness scenario by learning the common information that is shared between multiple related tasks and promotes sharing of model parameters to exploit the shared information across multiple tasks. The primary  issue  in  MTL setting is to appropriately learn the relation between the tasks \citep{happy2020apathy} otherwise can lead to negative performance. Several techniques such as grouped multiple tasks \citep{kang2011learning}, asymmetric MTL \citep{lee2016asymmetric}, multi-linear relationship networks \citep{long2017learning}, class relationship  \citep{wu2014exploring}, joint dynamic weighted \citep{das2018mitigating} has been proposed in the literature. From the literature it can be concluded that enforcing MTL to a scenario is challenging as it depends on the task behavior and the context of MTL based document analysis has not yet been much explored which encourages us to explore it in detail.

	\section{Dataset}
	\label{data_set}
		We have used \emph{L3iTextCopies} data-set \citep{Eskenazi2015}. This data-set is consisting of clean, text-only, typewritten documents which has $22$ actual pages. These pages has following characteristics: $1$ page of a scientific article with a single column header and a double column
		body, $3$ pages of scientific articles with a double column layout, $2$ pages of programming code with a single column layout, $4$ pages of a novel with a single column layout, $2$ pages of legal texts with a single column layout, $4$ pages of invoices with a single column layout, $4$ pages of payslips with a single column layout, $2$ pages of birth extract with a single column layout. Several variants of these $22$ pages are created by combining $6$ fonts: \emph{Arial}, \emph{Calibri}, \emph{Courier}, \emph{Times New Roman}, \emph{Trebuchet}, \emph{Verdana}; $3$ font sizes: $8$, $10$ and $12$ points; $4$ emphasis: \emph{normal}, \emph{bold}, and the combination of \emph{bold} and \emph{italic} which makes the total data-set size of $1584$ documents. Then these documents were printed by three printers (\emph{Konica Minolta Bizhub 223}, \emph{Sharp MX M904} and \emph{Sharp MX M850}) then these ones were scanned by three scanners at different resolutions between $150~dpi$, $300~dpi$ and $600~dpi$. Which finally generates a complete data-set of total $42768$ document images. 
		To obtain the word images, we apply \emph{Tesseract} OCR\footnote{\url{https://github.com/tesseract-ocr/}} to detect the word boundaries and then these ones are cropped from all the document images. To avoid the noisy elements, we only have considered the word images, more than $15 \times 15$ pixels in dimension. Whereas, to get the patches from a whole image, we crop patches of window size (standard input image size of ResNet) $224\times224$ pixels by sliding the window by $112 \times 112$ in horizontal and vertical directions.
	\vspace{-2mm}
	\section{Proposed Methods}
	\label{proposed_method}
	
	In this section, we have explained the architectures of our proposed technique for document attributes classification.
%
	As the base architecture, we propose a \emph{ResNet50} \citep{He2016} based model to perform ``single task learning'' i.e. ``Font Emphasis'', ``Font Type'', ``Font Size'', ``Scanning Resolution'' tasks separately and independently
	(see Figure~\ref{fig:archi_resnet_single}). A pre-trained (trained on \emph{ImageNet} data-set) \emph{ResNet50} model is used here and input images normalized in the same way, i.e. mini-batches of $3$-channel RGB images of shape ($3 \times H \times W$), where $H$ and $W$ are taken as $224$. The images are loaded in a range of $[0, 1]$ and then normalized using \emph{mean} = $[0.485, 0.456, 0.406]$ and \emph{std} = $[0.229, 0.224, 0.225]$. Hence, we obtain a pre-trained feature of size $2048$ (see Table.~1 in \cite{He2016}) from \emph{conv5\_x} layer. 
	
	After obtaining the $2048$ features, we add two subsequent fully connected layers : \fbox{ \strut \textbf{FC} ($2048 \rightarrow 512$)} (notation like this represents that it is a fully connected layer which has $2048$ input nodes and $512$ output nodes) and \fbox{ \strut \textbf{FC} ($512 \rightarrow 256$)}. Then, we add a \emph{batch normalization (BN)} layer which is followed by: \fbox{ \strut \textbf{FC} ($256 \rightarrow 256$)}. Then we again add another BN layer which is finally connected to one output head (among $4$ individual output heads). These heads are dedicated for each individual document attribute related tasks i.e. \emph{Font Emphasis}, \emph{Font Type}, \emph{Font Size} and \emph{Scanning Resolution}. 
	Each of these heads takes an input of $256$ values and provides an output of $\gamma$ values; where $\gamma =4$ for \emph{Font Emphasis} task, $\gamma =6$ for \emph{Font Type} task, $\gamma =3$ for \emph{Font size} task and $\gamma =3$ for \emph{Scanning Resolution} task (see Figure~\ref{fig:archi_resnet_single}). We train and test this network with either segmented word images or the patches and try to classify either each word images or each patches.
	
	\subsection{Multi Task Learning (MTL)}
	\label{MTL}
	As the first MTL architecture,
	we propose a multi task learning network where these four tasks can be performed using a single network (see Fig.~\ref{fig:archi_resnet_multiple}). The only difference of this architecture, compared to the previous one is that here last FC layer is finally connected to four individual heads (instead of one head in the previous network, shown in Fig.\ref{fig:archi_resnet_single}). 
	
	\subsection{Multi Task and Multi Instance (MI) Learning}
	\label{MI_MTL}
\begin{figure}[!htb]
		\centering{\includegraphics[trim = 8.2cm 13.6cm 5.4cm 5.6cm, clip,  scale = 0.42]{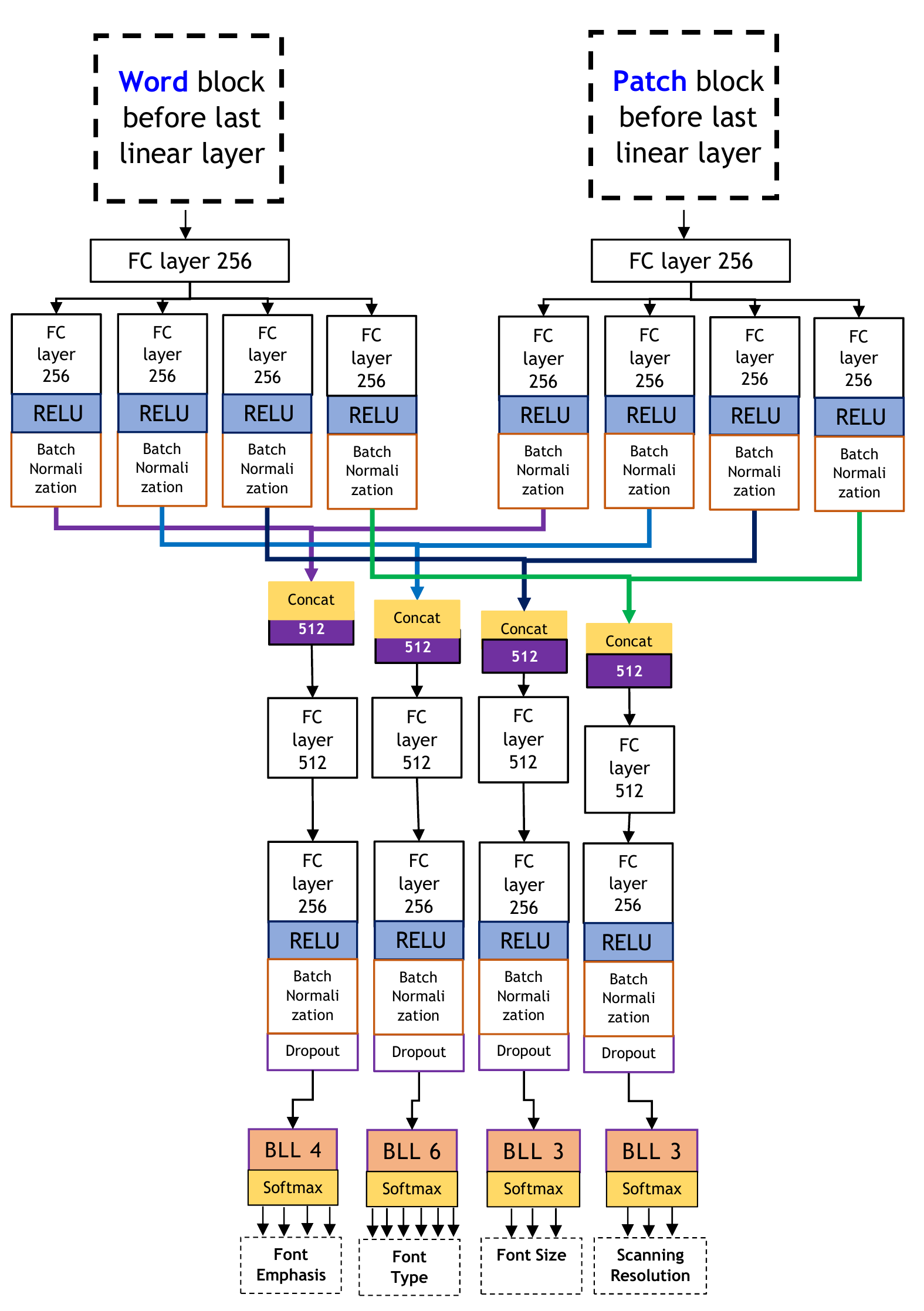}}
		\caption{The MTL + MI based architecture
		}
		\label{fig:archi_resnet_combine}
	\end{figure}

	As the second MTL based network, we propose to perform a linear combination of the segmented word based MTL network with the segmented patch based MTL network (see Fig.~\ref{fig:archi_resnet_combine}). 
	In general, a \emph{MTL+MI} based methods leverage the information which can be learned from other related tasks and it learns a general representation from all the available tasks.
	\begin{figure}[!htb]	       
		\begin{subfigure}[b]{0.49\linewidth}
			\centering{\includegraphics[trim = 7.7cm 8.6cm 23.6cm 3.2cm, clip,  scale = 0.42]{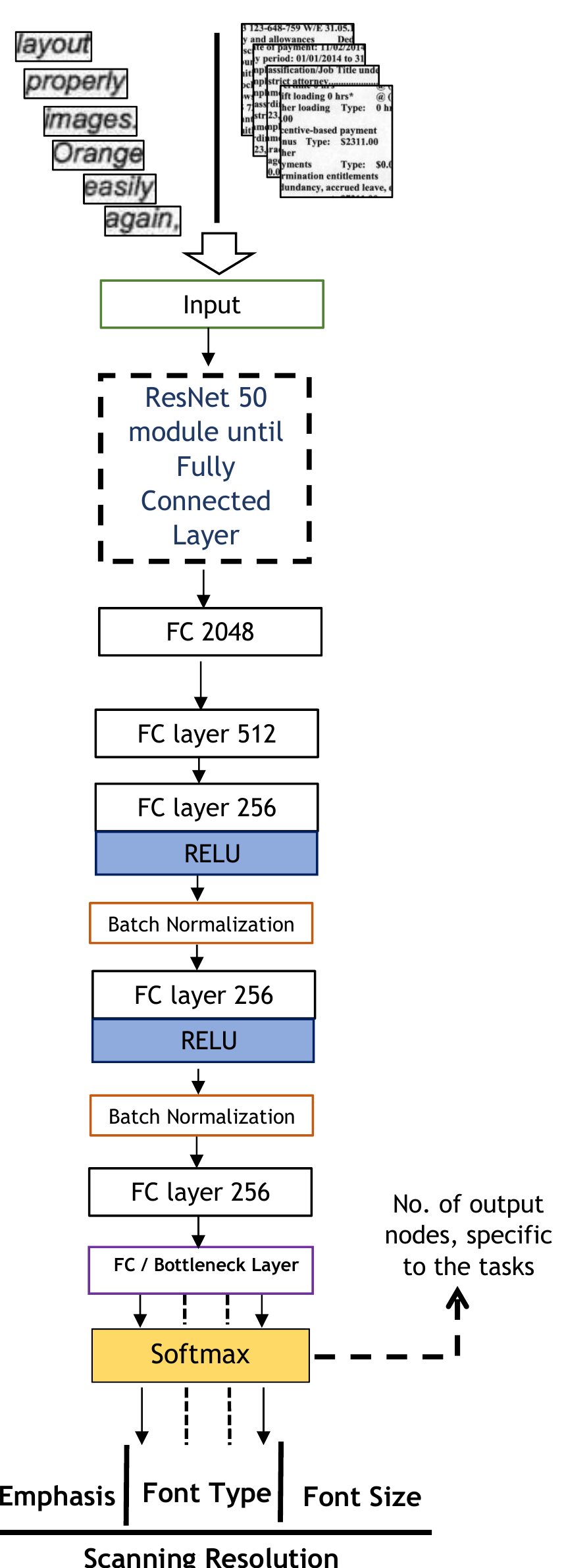}}
			\caption{}
			\label{fig:archi_resnet_single}
		\end{subfigure}
		\begin{subfigure}[b]{0.49\linewidth}
			\centering{\includegraphics[trim = 6.7cm 7.9cm 21.2cm 3.1cm, clip,  scale = 0.42]{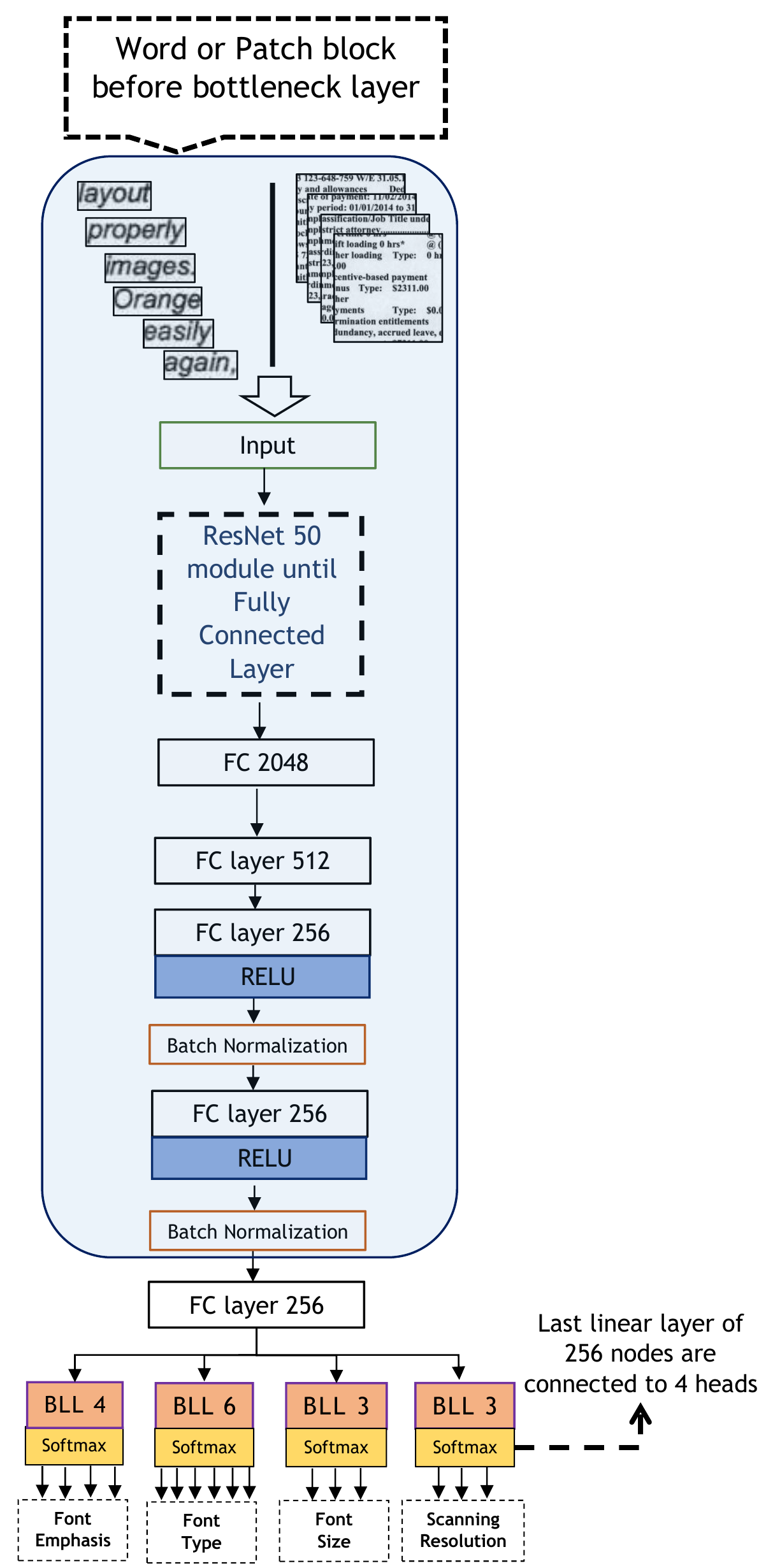}}
			\caption{}
			\label{fig:archi_resnet_multiple}
		\end{subfigure}	
		\caption{(a) The ``Single Tasks Learning'' based architecture. (b)The ``Multi Tasks Learning''  based architecture for document attribute classification.
		}
		\label{fig:archi_resnet}
	\end{figure}

	To combine two \emph{MTL based networks} i.e. \emph{MTL network} for word images and \emph{MTL network} for patches, the proposed architecture (see Fig.~\ref{fig:archi_resnet_combine}) remains same until last BN layer of the previous \emph{MTL network} (this block or portion of network is named as \emph{``Word or Patch block before bottleneck layer''} in Fig.~\ref{fig:archi_resnet_multiple}). The last \textbf{BN} layer has $256$ output nodes which are connected to another FC layer of $256$ nodes for both the word and patch level networks. Now the block of $256$ output nodes (for both the patch and word networks) are copied into $4$ heads (i.e. FC layer of word network as well as the patch network are separately connected to $4$ heads), where each head is consisting of \fbox{ \strut \textbf{FC} ($256 \rightarrow 256$)-\textbf{RELU}-\textbf{BN}}. Then the $1^{st}$ head of the word network is concatenated to the $1^{st}$ head of patch network. Same operation is performed for $2^{nd}$, $3^{rd}$ and $4^{th}$ heads. By each concatenation operation, we concatenate the $256$ nodes of word network and $256$ nodes of patch network to obtain total $512$ nodes. Now each of these $4$ group of concatenated $512$ nodes are connected to $4$ individual and independent head of \fbox{ \strut \textbf{FC} ($512 \rightarrow 512$)}.  Which are then finally connected to $4$ individual and independent blocks of \fbox{ \strut \textbf{FC} ($512 \rightarrow 256$)-\textbf{RELU}-\textbf{BN}-\textbf{DropOut}}.
	Finally, each head is connected to a \fbox{ \strut \textbf{FC} ($256 \rightarrow \gamma $)}, followed by \textbf{SoftMax} activation, where $\gamma$ represents the number of output nodes, dedicated to each individual document attributes classifications i.e. for \emph{Font Emphasis}, \emph{Font Type}, \emph{Font Size} and \emph{Scanning Resolution} tasks. The objective of this combined architecture is to get benefited from multi instance learning i.e. to train by using both the word and patch images together. Hence, the architecture can get benefited from the equal participation of both the words and patch features together.   
	
	\subsection{Dynamic Weighed Multi Task \& Multi Instance Learning}
	\label{Weighted_MI_MTL}
\begin{figure}[!htb]
		\centering{\includegraphics[trim = 2.2cm 10.6cm 4.8cm 3.2cm, clip,  scale = 0.40]{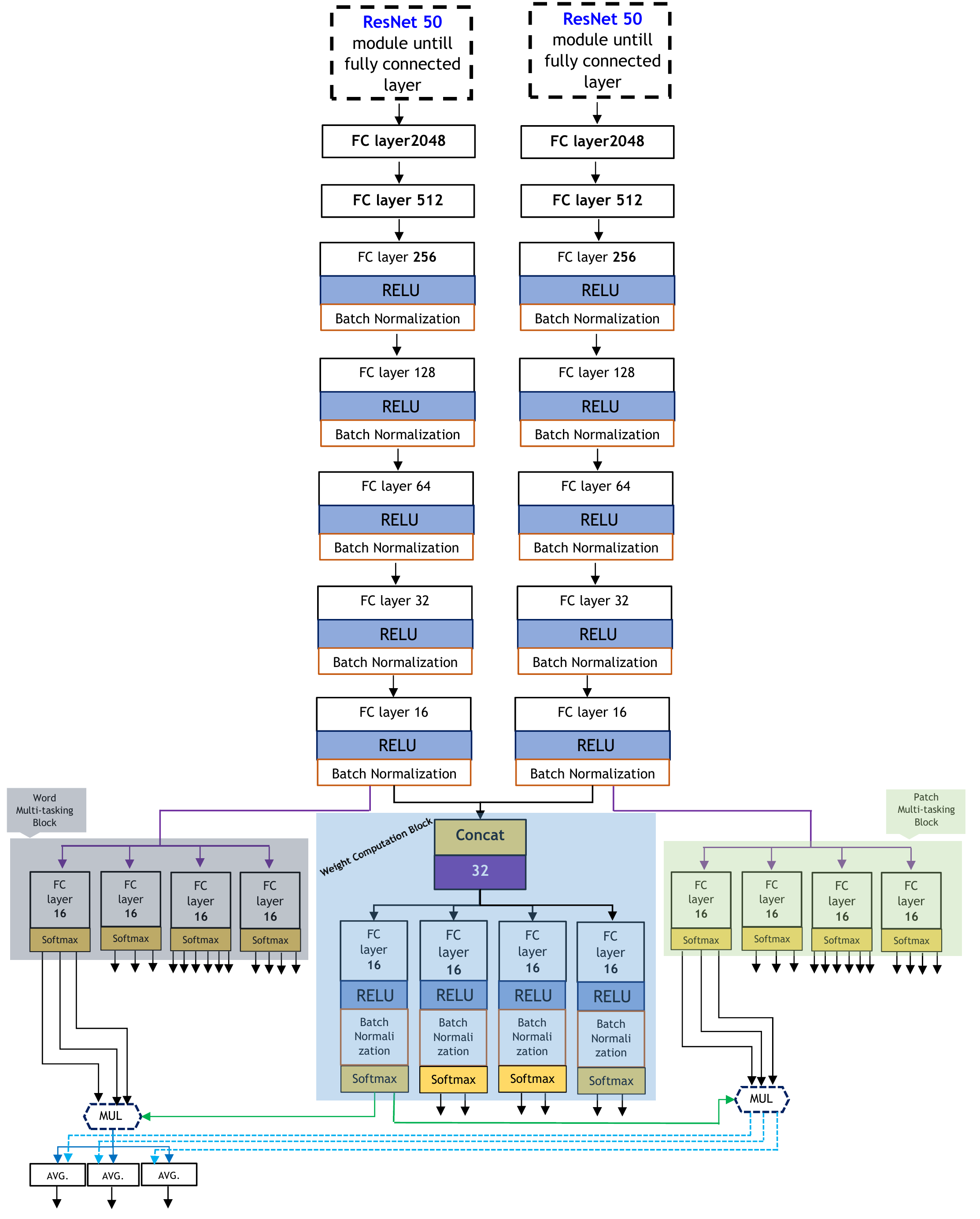}}
		\caption{The weighted MTL + MI based architecture
		}
		\label{fig:archi_resnet_combine_weighted}
	\end{figure}

	In the previous architecture (see Section~\ref{MI_MTL}), we have applied equal weights (or equal participation) to both the words and patch features. But it could also be possible that due to this equal weighted fusion, instead of improving, the accuracy may get decreased. The most probable reason of this kind of problem is the unequal influence of one instance (either words or patches), compared to other. Hence, to avoid this problem, we propose to perform multi instance learning, based on automatically calculated weights which are applied to both the instances. The architecture is shown in Fig.~\ref{fig:archi_resnet_combine_weighted}.  	
%
	
	After obtaining $2048$ pre-trained features from word and patch images, we pass them through \fbox{ \strut \textbf{FC} ($2048 \rightarrow 512$)}.
	Then each of these feature set ($512$ features) is passed through several FC layers like : 
	\fbox{ \strut \textbf{FC} ($512 \rightarrow 256$)-\textbf{RELU}-\textbf{BN}}; \fbox{ \strut \textbf{FC} ($256 \rightarrow 128$)-\textbf{RELU}-\textbf{BN}};~\fbox{ \strut \textbf{FC} ($128 \rightarrow 64$)-\textbf{RELU}-\textbf{BN}}; \fbox{ \strut \textbf{FC} ($64 \rightarrow 32$)-\textbf{RELU}-\textbf{BN}};~\fbox{ \strut \textbf{FC} ($32 \rightarrow 16$)-\textbf{RELU}-\textbf{BN}}; Then $16$ output features of word (as well as patch) images are get connected to $4$ output heads (see ``Word Multi-tasking block'' at the left and ``Patch Multi-tasking block'' at the right of Fig.~\ref{fig:archi_resnet_combine_weighted}) 
	The weights are automatically computed by initially combining $16$ features of ``word'' network and $16$ features of ``patch'' network together. Then these $32$ combined features are individually connected to $4$ heads of \fbox{ \strut \textbf{FC} ($32 \rightarrow 16$)-\textbf{RELU}-\textbf{BN}-\textbf{SoftMax}}; As it is visible from the Fig.~\ref{fig:archi_resnet_combine_weighted} that each output head is outputting $2$ weight values. The $1^{st}$ weight value is dedicated for the output of ``word multi-tasking block'' and the $2^{nd}$ weight value is dedicated for ``patch multi-tasking block''. 

	After obtaining two weight values from each output heads, 
	these ones are multiplied with the outputs of each output heads of ``word multi-tasking block'' as well as ``patch multi-tasking block''. For example, the $3$ outputs of $1^{st}$ output head (dedicated for ``Font Emphasis'' task) of ``word multi-tasking block'' are get multiplied with the $1^{st}$ weight value, coming from the $1^{st}$ head of ``weight computation block''. In the same manner, the $2^{nd}$ weight from the  $1^{st}$ head of ``weight computation block'' got multiplied with the $3$ outputs of $1^{st}$ output head (dedicated for ``Font Emphasis'' task) of ``patch multi-tasking block''. Finally, the pair of $3$ output values are element-wise averaged to generate $3$ final output values, dedicated for ``font emphasis'' task. In the same manner, we perform the weighted average of other three tasks i.e. \emph{Font Type}, \emph{Font Size} and \emph{Scanning Resolution} tasks to generate the final averaged outputs.
	
	 
	\subsection{Calculation of Loss in Multi Task \& Multi Instance Learning}
	In this section, we will discuss about the loss function of all the above defined $4$ different architectures. \\
	\textbf{a) Single Task Learning Loss}: In the case of \textit{Single Task Learning} (mentioned in Section~\ref{proposed_method}), the cross entropy loss $\mathcal{L}_{STL}$ is defined as follows :
	
	\begin{equation}
	   \mathcal{L}_{STL}(\mathbf{X};\Theta) = \sum_{k=1}^{K} - y_k ~log P(y_k |\mathbf{X}, \Theta)
	   \label{eq:cross_entropy} 
	\end{equation}
	where $K$ is the total number of output classes, $\mathbf{X}$ and $\Theta$ are the input and parameters of the network, $y_k$ is the true label of input $\mathbf{X}$ and
	$P(y_k|\mathbf{X},\Theta)$ is the predicted probability of class $k$ for the given image $\mathbf{X}$ and parameter $\Theta$. As mentioned in Section~\ref{proposed_method} that the input could be either segmented word images or patch images whereas the network could be trained for the classification of any of these $4$ attributes i.e. for \emph{Font Emphasis}, \emph{Font Type}, \emph{Font Size} and \emph{Scanning Resolution} tasks.
	
	\textbf{b) Multi-Task Learning Loss}: Compared to single-task learning, multi-task learning can gain better performance by jointly learning different tasks. Multi-task learning is an optimization problem for multiple objectives. The loss function of multi-task learning (mentioned in Section~\ref{MTL}) is calculated by summing up the different tasks in the following manner.
	  
	\begin{equation}
	  \mathcal{L}_{MTL}(\mathbf{X};\Theta) = \sum_{t=1}^{T} \mathcal{L}_t(\mathbf{X};\Theta_t)
	  \label{eq:loss_MTL} 
	\end{equation}
	where $T$ is the number of supervised tasks for which we would like to train in the \textit{MTL} mode, where the loss of each task $\mathcal{L}_t(\mathbf{X};\Theta_t)$ is calculated by using same formula as in Equation~\ref{eq:cross_entropy}, $X$ is input of the model and $\Theta=\{\Theta_t\}$ are the parameters of the portion of the network, related to each tasks. In our case, we have $4$ independent tasks i.e. \emph{Font Emphasis}, \emph{Font Type}, \emph{Font Size} and \emph{Scanning Resolution} tasks, hence $T=4$. 
	
	 \textbf{c) Multi-Task and Multi-Instance Learning Loss}: For the case of \textit{Multi Instance} and \textit{Multi-Task} learning (see Section~\ref{MI_MTL}), the loss function is calculated in the following manner. Each word image sample, denoted by $\mathbf{X}_i^{word}$ and the patch image sample, denoted by $\mathbf{X}_i^{patch}$, we have $T$ number of label information, where $T$ is the total number of existing tasks. Thus all the tasks, learn together their features in a joint feature space $f\in \mathcal{F}$ through learning the weights for individual tasks. The learned features, before the concatenation operation in Fig.~\ref{fig:archi_resnet_combine} can be represented by $\mathcal{Z}_t^{{word}_{256}}$, $\mathcal{Z}_t^{{patch}_{256}}$ as follows: 
	\begin{equation}
	 \resizebox{.85\hsize}{!}{
		$ \mathcal{Z}_t^{{word}_{256}} = \mathcal{F}(\mathbf{X}^{word}, \Theta^{word}_t); ~~~~~  \mathcal{Z}_t^{{patch}_{256}} = \mathcal{F}(\mathbf{X}^{patch}, \Theta^{patch}_t)$
		}
	\end{equation}
	where $\Theta^{word}_t$ and $\Theta^{patch}_t$ are the parameters of the word branch and patch branch of each task before the concatenation. Then these feature vectors i.e.  $\mathcal{Z}_t^{{word}_{256}}$, $\mathcal{Z}_t^{{patch}_{256}}$ are concatenated in the following manner:  
	
	\begin{equation}
		\mathcal{Z}_t^{{concat}_{512}} = Concat(\mathcal{Z}_t^{{word}_{256}}, \mathcal{Z}_t^{{patch}_{256}})
	\end{equation}
	
	where $t \in {1,..,T}$.  
	After that the \textit{MTL} loss function is calculated in the same manner as : 
	\begin{equation}
		\mathcal{L}_{MTL+MI}^{concat}([\mathbf{X}^{word}, \mathbf{X}^{patch}];\Theta) = \sum_{t=1}^{T} \mathcal{L}_t(\mathcal{Z}_t^{{concat}_{512}} ;\Theta_t)
		\label{eq:loss_MTL_combined} 
	\end{equation}
 where $\Theta_t$ is the parameter of the branch of each task after the concatenation of features. \\
 \textbf{d) Dynamic  Weighed  Multi-Task  and  Multi-Instance Learning Loss:} Rather than applying equal weights to both the words and patch features, the dynamic weighted version of \textit{Multi Instance} and \textit{Multi-Task} learning considers the different contribution of learned words and patch features by weighting the output of word instance $y_k^{{word}_{t}}$ and the output of patch instance $y_k^{{patch}_{t}}$ to formulate the final output value $\mathcal{Y}_{t,k}^{{Avg}}$ as described in Section~\ref{Weighted_MI_MTL}:
 \begin{equation*}
	 	\mathcal{Y}_{t, k}^{{Avg}} = Elementwise\_Avg(w^1_t(\Psi) \times ~y_k^{{word}_{t}}, w^2_t (\Psi)\times ~ y_k^{{patch}_{t}})
 \end{equation*}
 where ${w^1_t, w^2_t}$ are the weights, associated to the output of each instances i.e. the words ($y_k^{{word}_{t}}$) and patch ($y_k^{{patch}_{t}}$) respectively. These weights are calculated dynamically during training process. The dynamic weights i.e. $\{\alpha_i=w_i(\Psi)\}$ are learned automatically by the dynamic weights learning module. Particularly, since the dynamic task weights $\{\alpha_i\}$ are the outputs of the softmax layer, $\sum\alpha_i=1$.  $\Psi$ are the parameters of the dynamic weights learning module. Note that $\Psi \not\subset \Theta$. The parameters of the dynamic weights learning module $\Psi$ and the parameters of the network $\Theta$ are optimized simultaneously with the total loss $\mathcal{L}_{MTL+MI}^{weighted}$ given by:
  \begin{equation}
	\mathcal{L}_{MTL+MI}^{weighted} = \sum_{t=1}^{T} \mathcal{L}_{MTL+MI}^{t}([\mathbf{X}^{word}, \mathbf{X}^{patch}];\Theta)
	\label{eq:loss_MTL_weighted} 
 \end{equation}

 where $\mathcal{L}^t_{MTL+MI}$ is the loss of each task with the dynamic weighted output $\mathcal{Y}_{t,k}^{{Avg}}$: 
 \vspace{-3mm}
 \begin{equation}
 \resizebox{.85\hsize}{!}{
 	$
 	\mathcal{L}^t_{MTL+MI}([\mathbf{X}^{word}, \mathbf{X}^{patch}];\Theta) = \sum_{k=1}^{K} - \mathcal{Y}_{t,k}^{Avg} ~log P(\mathcal{Y}_{t,k}^{Avg} |\mathbf{X}, \Theta)$
 }
 \end{equation}

\subsection{Component based majority voting for entire document image classification}
\label{voting_scheme}
\begin{figure}[h!]
		\centering{\includegraphics[trim = 1.8cm 19.1cm 9.2cm 3.8cm, clip,  scale = 0.83]{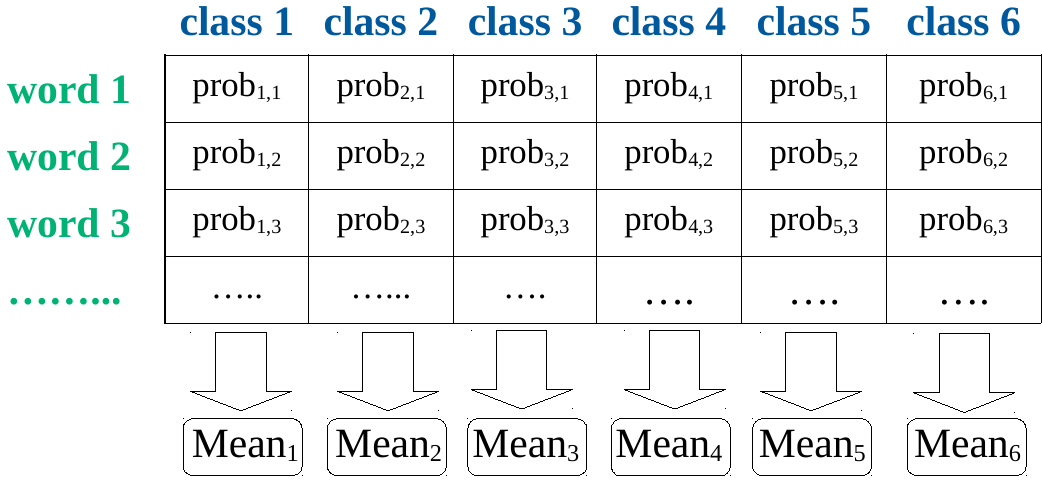}}
		\caption{The linear combination of multi-tasking Res-Net of word images and multi-tasking Res-Net of patch images.
		}
		\label{fig:font_probability}
	\end{figure}

	The training, validation and testing are performed by using either word images or the patch images. The classification of an entire document image is done by considering a voting mechanism, where the posterior probabilities of all the word image (or patches) are taken into account to compute the mean posterior probability of each document page. Such a mechanism is shown in Fig.~\ref{fig:font_probability}, where the word images, obtained from each page are arranged row wise and classes are arranged along columns. The posterior probabilities of each word to belong into a specific class is noted in each cells. Hence, by taking column wise mean followed by the maximum of these mean values, we can decide the class of complete document image. The same principal is applicable by considering the patches for the classification of complete document image.
	Let's say $P(y_k^i|\mathbf{X}_i^p,\Theta)$ is the probability to belong in $k^{th}$ (here in this example, shown in Fig.~\ref{fig:font_probability}, $k \in {1,...,6}$) class for the $i^{th}$ word, taken from $p^{th}$ document page of the data-set.
		
	The average probability of all the words (belongs to $p^{th}$ document page) for each class is calculated by:
	\begin{equation}
		\bar{\mathcal{P}} = \frac{1}{n}\left(\sum_{i=1}^{n}P(y_k^i|\mathbf{X}_i^p\right); ~ i \in {1,..,n}
	\label{eq:mean_prob} 
	\end{equation}
	where there are $n$ number of words exists in $p^{th}$ document page. Now the document page is belongs to which class is decided by computing the maximum of $\bar{\mathcal{P}}$ in the following manner :
	\begin{equation}
	 \mathcal{P}_{max}  = max \left( \bar{\mathcal{P}} \right) 
	 \label{eq:max_prob} 
	\end{equation}
	Thanks to our proposed voting mechanism, even if the predicted posterior probabilities of one or multiple components for the correct class is weaker than the other classes, thanks to the mean based voting scheme, the high posterior probabilities of the remaining components of full page image will compensate and finally the mean accuracy of the true class will be superior.    
	 
	\section{Experiments}
	\label{experiments}
	In this section, we explain our experimental setting and the results obtained are discussed with detailed analysis.
	The implementation details of all the above mentioned networks are mentioned in  Section~\ref{implement_details} of supplementary materials.

	\subsection{Results and discussion}
	\label{results_discussion}
	In this section, we have discussed the results of different aforementioned proposed networks. As mentioned in Section~\ref{data_set} that out of $42,768$ images, $70\%$ images ($29,950$) are taken for training and $10\%$ images ($4278$) are taken for validation. Hence, by considering only valid word images (see Section~\ref{data_set}), the training data set is consisting of $8,099,561$ images whereas the validation set is consisting of $1,344,016$ word images. Whereas, by considering only valid patches, the training data set is consisting of $7,261,574$ images whereas the validation set is consisting of $1,309,966$ word images.  
	Furthermore, the test set is consisting of $20\%$ of the total images i.e. $8557$ images from which we could extract $5,990,78$ word image and $1,324,58$ patch images. 
	
	\begin{table}[h!]
		\small{
		\caption{Testing accuracy of STL \& MTL based network for word and patch level images }
		\centering
		\begin{tabular}{|P{1.6cm}|P{0.8cm} P{0.8cm} P{1.5cm} P{1.7cm}|}
			\hline
			& \multicolumn{4}{c|}{\textbf{Top-1 accuracy}} \\ \hline
			& \textbf{Font Type} & \textbf{Font Size} & \textbf{Font Emphasis} & \textbf{Scanning Resolution} \\ \hline
			STL Word    & 0.8508     &  0.9095    &    0.9413     &  0.9933 \\ \hline
			STL Patch    &  \textbf{0.9452}     &  \textbf{0.9661}   &   \textbf{0.9780}     &  \textbf{0.9963 }  \\ \hline \hline
			MTL Word  &   \textcolor{blue}{0.8887}    &   \textcolor{blue}{0.9137}  &   0.9379       &    \textcolor{blue}{0.9939} \\ \hline
			MTL Patch & \textcolor{blue}{\textbf{0.9512}}    &  \textcolor{blue}{\textbf{0.9760}}    &     \textcolor{blue}{\textbf{0.9827}}      &        \textcolor{blue}{\textbf{0.9965}}   \\ \hline
		\end{tabular}
		
		\label{STL_accuracy}
		}
	\end{table}
	
	\begin{table}[]
		\small{
		\caption{Training (gray colored row), validation (light blue colored row) and testing (orange colored row) accuracies of MTL based network for word and patch level images }
		\centering
		\begin{tabular}{|P{2cm}|P{0.8cm} P{0.8cm} P{1.5cm} P{1.7cm}|}
			\hline
			& \textbf{Font Type} & \textbf{Font Size} & \textbf{Font Emphasis} & \textbf{Scanning Resolution} \\ \hline
			\multirow{3}{*}{\shortstack{\textbf{Word} Level \\ with multiple \\ \textbf{FC} layers}} & \mycc 0.9703     &  \mycc 0.8917    &    \mycc 0.9557     &  \mycc 0.9828  \\
			& \mybb 0.6073     & \mybb 0.6449    &   \mybb 0.7042     & \mybb 0.8431 \\
			& \myrr  0.6986    & \myrr  0.6933   &   \myrr  0.8359    & \myrr  0.9386\\
			\hline
			
			\multirow{3}{*}{\shortstack{\textbf{Word} Level \\ with \textbf{AlexNet} \\ layers}} & \mycc 0.4094     &  \mycc 0.4199    &    \mycc 0.5860     &  \mycc 0.8642  \\
			& \mybb 0.3447     & \mybb 0.4013    &   \mybb 0.5203     & \mybb 0.7677 \\
			& \myrr   0.3962   & \myrr 0.3907   &   \myrr 0.5992    & \myrr 0.8328 \\
			\hline
			\hline
			

			\multirow{3}{*}{\shortstack{\textbf{Patch} Level \\ with multiple \\ \textbf{FC} layers}} & \mycc 0.8631     &  \mycc 0.8999    &    \mycc 0.9131     &  \mycc 0.9933  \\
			& \mybb 0.7751     & \mybb 0.8544    &   \mybb 0.9501     & \mybb 0.9979 \\
			& \myrr 0.7777     & \myrr 0.8563    &   \myrr 0.9421     & \myrr 0.9899 \\
			\hline
			\multirow{3}{*}{\shortstack{\textbf{Patch} Level \\ with \textbf{AlexNet} \\ layers}} & \mycc 0.3688     &  \mycc 0.5932    &    \mycc 0.6368     &  \mycc 0.9444  \\
			& \mybb 0.3802     & \mybb 0.6167    &   \mybb 0.7870     & \mybb 0.9806 \\
			& \myrr  0.3600   & \myrr 0.5957  &   \myrr 0.8074   & \myrr 0.9562\\
			\hline
			
		\end{tabular}
		\label{MTL_accuracy_1}
		}
	\end{table}

	\subsubsection{Accuracies of STL \& MTL based networks}
\begin{figure}[!htb]
		\centering
		\begin{minipage}{\linewidth}
			\centering{\includegraphics[trim = 4.2cm 1.6cm 3.4cm 1.3cm, clip,  scale = 0.21]{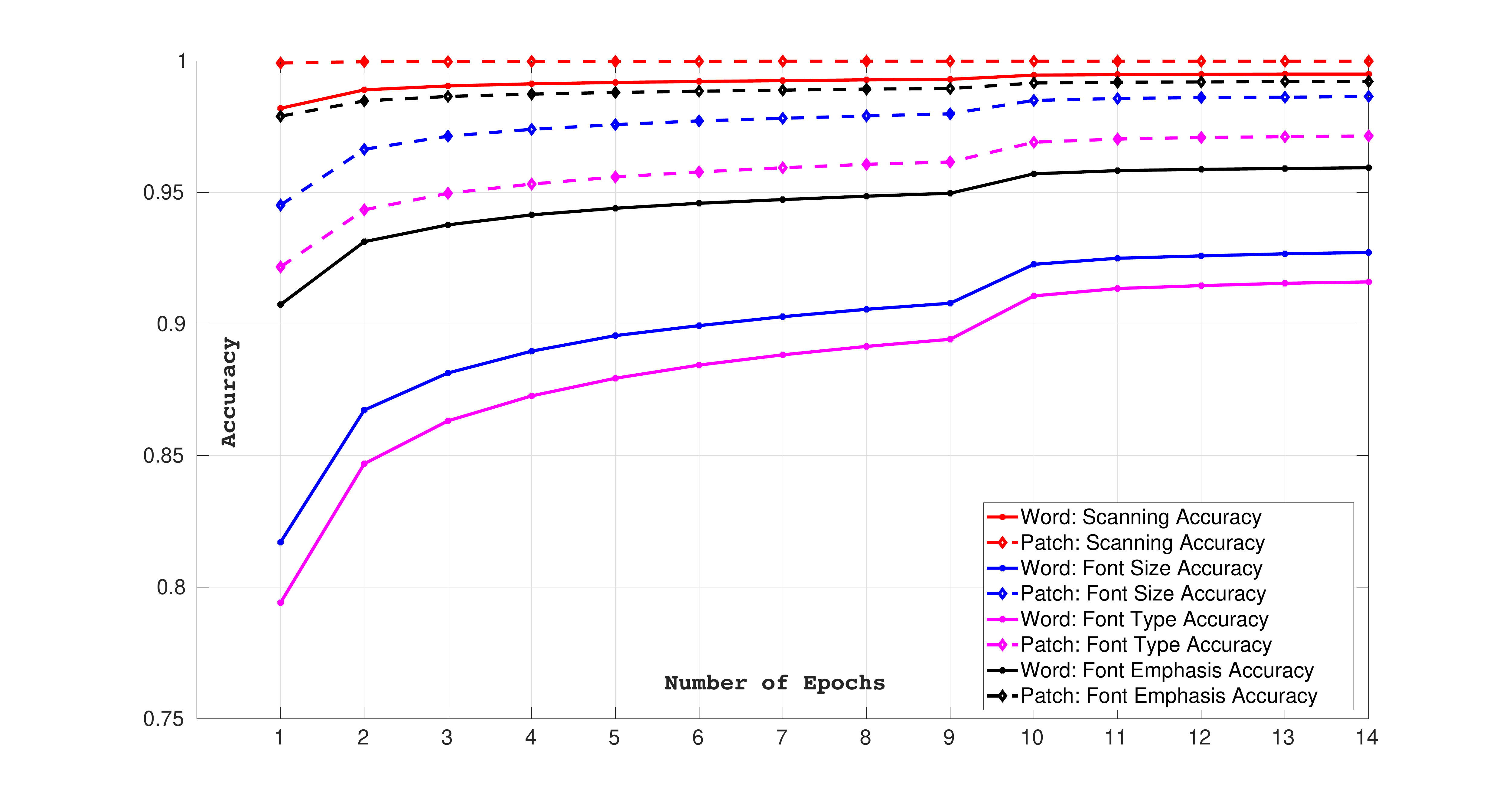}}
			\\{\hspace{5mm}\small{(a)}}
		\end{minipage}\\
		\caption{Training accuracies of word based and and patch based MTL networks for ``Font type'', ``Font size'', ``Font emphasis'' and ``Scanning resolution'' tasks.
		}
		\label{fig:multitask_patch_word_together}
	\end{figure}

	The testing accuracy of STL based network (see Section~\ref{proposed_method}) and MTL based network (see Section~\ref{MTL}) in shown in Table~\ref{STL_accuracy}, where it can be visible that the patch level accuracy is better than the word level for all the document attribute classifications i.e. for \emph{font type}, \emph{font size}, \emph{font emphasis} and \emph{scanning resolution}. 
	It can also be seen that in the case of word and patch level classification, the \textit{Multi-task learning} has performed better than \textit{Single-task learning}, which was expected. The training accuracies of MTL based word level as well as patch level networks are shown in Fig.~\ref{fig:multitask_patch_word_together} (in the same manner, the training and validation accuracies of MTL based word and patch level networks are shown in Fig.~\ref{fig:multitask_patch_word} and are discussed in Section~\ref{train_val_accura}). It can be visible that the accuracies of all the $4$ tasks increases with the iteration of each epochs and patch level network has performed better than word level network.     
	
	\subsubsection{Small data-set formation} 
	\label{small_dataset}
	To test the accuracy of following two architectures i.e. ``Multi-task and Multi-instance Learning'' and ``Weighted Multi-task and Multi-instance Learning'', we choose only a part of the complete data-set in the following manner to obtain faster results and to test various architectural modifications. 	
	The document images (also the cropped word and patch images) in the data-set are actually labeled (i.e. the ground truth) in one out of $216$ (by combining all the output classes of the four tasks i.e. $4 \times 6 \times 3 \times 3 = 216$) independent classes. We choose $400$ word \& patch images for training, $100$ word \& patch images for validation and $150$ word \& patch images for testing from each of the $216$ classes\footnote{if there are not enough number of word and/or patch images exists in any of these $216$ classes then we first count total number of word (say $c_{word}$) and patch (say $c_{patch}$) images in this class and then choose $c_{patch}$ numbers of word and patch images if $ c_{patch} < c_{word}$ or vice-versa.}. These limited number of images are chosen by calculating total number of foreground pixels (we have applied adaptive threshold based binarization using Otsu's method) in the image followed by choosing the images which has high number of foreground pixels. Hence in this manner, we choose $83,389$ word and $83,389$ patch images for training and $20,731$ word and $20,731$ patch images for validation.

	 \begin{table}[]
	 	\small{
	 	\caption{Training, validation and testing accuracies of MTL and MI based network by combining word and patch level images }
	 	\centering
	 	\begin{tabular}{|P{2cm}|P{0.8cm} P{0.8cm} P{1.5cm} P{1.7cm}|}
	 		\hline
	 		& \textbf{Font Type} & \textbf{Font Size} & \textbf{Font Emphasis} & \textbf{Scanning Resolution} \\ \hline
	 		\multirow{3}{*}{\shortstack{\textbf{Late} concat \\ \textbf{multiple} FC\\ layers}} & \mycc 0.9540     &  \mycc 0.9611    &    \mycc 0.9766     &  \mycc 0.9973  \\
	 		& \mybb 0.1859     & \mybb 0.4090    &   \mybb 0.6757     & \mybb 0.9334 \\
	 		& \myrr 0.1885    & \myrr  0.4165  &   \myrr 0.7217    & \myrr 0.7759 \\
	 		\hline
	 		\multirow{3}{*}{\shortstack{\textbf{Early} concat \\ \textbf{multiple} FC\\ layers}} & \mycc 0.9680     &  \mycc 0.9606    &    \mycc 0.9792     &  \mycc 0.9975  \\
	 		& \mybb 0.5287     & \mybb 0.5995    &   \mybb  0.6228     & \mybb 0.9740 \\
	 		& \myrr 0.4918     & \myrr 0.4805    &   \myrr 0.6318     & \myrr 0.8147 \\
	 		
	 		\hline
	 		\multirow{3}{*}{\shortstack{\textbf{Early} concat \\ \textbf{AlexNet} like \\ Conv. layers}} & \mycc 0.2929     &  \mycc 0.6037    &    \mycc 0.5895     &  \mycc 0.9754  \\
	 		& \mybb 0.2221     & \mybb 0.3756    &   \mybb 0.5120     & \mybb 0.9438 \\
	 		& \myrr 0.2197    & \myrr 0.3076    &   \myrr  0.4945    & \myrr 0.7998  \\
	 		\hline
	 		\hline
	 	\end{tabular}
	 	\label{MTL_accuracy_2}
	 	}
	 \end{table}	
	  
	\subsubsection{Accuracy of  Multi-task and Multi-instance Learning }
	\label{accu_multi_task_multi_instance}
	The first test is performed by using the architecture shown in Fig.~\ref{fig:archi_resnet_combine}, where the fusion of convoluted features from word images and patch images are performed lately (i.e. the $2048$ number of pre-trained ``ResNet-50'' are passed through several block of FC feed forward networks to finally generate $256$ number of features). The results of this network is shown in Table~\ref{MTL_accuracy_2} and are mentioned in first row as ``Late concat multiple FC layers''. Although, we have obtained high training accuracies for all the $4$ tasks but the validation and testing accuracies got drastically decreased for all the tasks (whereas the ``scanning resolution'' task shows comparatively better performance than other three tasks). This is a classical case of ``model over-fitting'' where the model is able to learn very well on training data but it fails to perform on validation and tes\textbf{}ting data. 
	
	We tried several approaches like increasing and decreasing the percentage of dropout nodes, completely removing all dropout layers, adding more dropout layers at every block of FC layers, removing all the batch normalization layers from the network etc. to overcome this problem. All these trials doesn't helped much and the problem of over fitting sustained. We suspected that the late concatenation of patch and word level features could be the possible reason for this over-fitting problem. Hence, we decided to concat the $2048$ number of pre-trained ``ResNet-50'' features at the early stage of the network. Hence, we designed another network, where $2048$ number of pre-trained ``ResNet-50'' features of word and patch images are concatenated at the early stage to generate $4096$ features. The architectural details of this network is mentioned in Section~\ref{early_concat_MTL_MI} of the supplementary material.
	
	By using this network, although the validation and testing accuracies has highly improved but still there remains a strong gap between the training accuracies (shows unconventionally very high) and testing/validation accuracies of the $4$ tasks.  
	By suspecting that the pipeline connection of several FC layers could be the possible reason of this gap between the training and testing/validation accuracies of the $4$ tasks, we tried to reduce the number of FC layers. This attempt doesn't helped much and the problem of ``model over fitting'' still remained.  Please look at Table~\ref{supp_MTL_accuracy_2} and corresponding discussion in supplementary materials for more details.
	Furthermore to handle this problem of model over-fitting, we proposed to replace the FC layers by convolutional layers. Here, we adopt the popular ``AlexNet'' \citep{Krizhevsky2017} like architecture instead of multiple FC layers, after obtaining $2048$ number of pre-trained ``ResNet-50'' features from word and patch images. We have named this network as ``Early concat AlexNet like conv. layers'' and it's results are shown in $3^{rd}$ row of Table~\ref{MTL_accuracy_2}. The architectural details of this network is mentioned in Section~\ref{alexnet_MTL_MI}. 
	
	It can be visible from the results, mentioned in Table~\ref{MTL_accuracy_2} that although ``Early concat AlexNet like conv. layers'' network is able to somehow overcome the problem of model over fitting and the difference between training and validation/testing accuracy got decreased (still for ``Font Size'' task, this difference remains significant), the overall accuracy of the network got significantly reduced. Instead of ``AlexNet'' like architecture, we have also tried ``VggNet'' \citep{He2016} like architecture in the same manner to see whether we can obtain higher accuracies and can also overcome the problem of model ``over-fitting''. This trial successfully overcome the problem of model ``over-fitting'' but the overall accuracy of the network also got significantly decreased. Please see Table.~\ref{supp_MTL_accuracy_2} and corresponding discussions in supplementary materials for more details. 
	
	\begin{table}[]
		\small{
		\caption{Accuracies of Dynamic weighted MTL and MI based network by combining word and patch level images }
		\centering
		\begin{tabular}{|P{2.2cm}|P{0.8cm} P{0.8cm} P{1.5cm} P{1.7cm}|}
			\hline
			& \textbf{Font Type} & \textbf{Font Size} & \textbf{Font Emphasis} & \textbf{Scanning Resolution} \\ \hline	
			\hline
			\multirow{3}{*}{ \shortstack{\textbf{Late} concat \\ \textbf{multiple} FC \\ layers} } & \mycc 0.5265     &  \mycc 0.7786    &    \mycc 0.8860     &  \mycc 0.9866  \\
			& \mybb 0.2976     & \mybb 0.4196    &   \mybb 0.4683     & \mybb 0.7484 \\
			& \myrr 0.2898    & \myrr 0.3794   &   \myrr 0.4544     & \myrr 0.7000  \\
			\hline
			\multirow{3}{*}{ \shortstack{\textbf{Late} concat \\ \textbf{AlexNet} like \\ Conv. layers} } & \mycc 0.2854     &  \mycc 0.5113    &    \mycc 0.5501     &  \mycc 0.9427  \\
			& \mybb 0.2114     & \mybb 0.3572    &   \mybb 0.3575     & \mybb 0.8673 \\
			& \myrr 0.2038   & \myrr 0.3190   &   \myrr  0.3454   & \myrr 0.7679 \\
			\hline
		\end{tabular}
		\label{MTL_weightedaccuracy}
		}
	\end{table}

	\subsubsection{Accuracy of Dynamic Weighted Multi-task and Multi-instance Learning }
	\label{weighted_accu_multi_task_multi_instance}
	 The network is shown in Fig.~\ref{fig:archi_resnet_combine_weighted} and the corresponding details about this architecture are mentioned in Section~\ref{Weighted_MI_MTL}. It can be visible from the accuracies, mentioned at the $1^{st}$ row in Table~\ref{MTL_weightedaccuracy} that the accuracies are not get further improved, compared to the ``Multi-task and Multi-instance Learning'' architectures and the problem of ``model over-fitting'' still remains. Furthermore, to handle the problem of ``model over-fitting'', we tried to adopt ``AlexNet'' like architecture by replacing the FC layers with several convolution layers. The details of this network is mentioned in Section~\ref{alex_net_weighted_multi_task_multi_instance} in the supplementary materials. The results of this network is shown in $2^{nd}$ row of Table~\ref{MTL_weightedaccuracy}. It can be visible from the results that the problem of model over fitting are seems to get resolved for ``Font Type'' and ``Scanning Resolution'' tasks but for other two tasks i.e. for ``Font Size'' and ``Font Emphasis'' tasks, the model over fitting problem persists. More importantly, the overall accuracies of these $4$ tasks become lesser than the ``Late concat multiple FC layers'' model.
	 
	 One interesting fact can be noticed from Table.\ref{MTL_accuracy_1} that even when we are using only word level network (see the $1^{st}$ row of Table.\ref{MTL_accuracy_1}) on the smaller dataset (see Section.~\ref{small_dataset}), the issue of ``model over fitting'' persists. This problem of ``model over fitting'' can be overcome by using ``AlexNet'' like layers instead of pipe lined FC layers. The results of this network, named as ``Word Level with AlexNet layers'' is shown at the $2^{nd}$ row of Table.\ref{MTL_accuracy_1}, where we can see the difference between the training and validation/testing accuracies for these $4$ tasks got reduced. But the overall accuracies of all the $4$ tasks got decreased than the ``Only Word Level'' network. 
	  
	 The results of patch level network (denoted as ``Patch level with multiple FC layers'') on small dataset, the problem of ``model over fitting'' is not visible (see the $3^{rd}$ row of Table.\ref{MTL_accuracy_1}). Moreover, the accuracies of all the $4$ tasks are quite higher compared with it's counter part i.e. ``Word level with multiple FC layers'' network. We further tested the performance of ``AlexNet'' like layers instead of pipe lined FC layers and this network is named as ``Patch level with AlexNet layers''. It can be visible from the results, (see the $4^{th}$ row of Table.\ref{MTL_accuracy_1}) that the problem of ``model over fitting'' is also disappeared in this case also but the accuracy got decreased, compared to ``Patch level with multiple FC layers'' network. 
	 
	 Among our various experiments with ``MTL and MI'' based network which combines word and patch level images, the ``Early concat multiple FC layers'' network performed better than others. Whereas, the experiments with various weighted ``MTL and MI'' based networks, the ``Late concat multiple FC layers'' network performed better. But the accuracies of both these networks i.e. ``Early concat multiple FC layers'' and ``Late concat multiple FC layers'' couldn't outperform ``Patch Level with multiple FC layers'' network. Hence, from all these above mentioned experiments, we can conclude that ``Patch Level with multiple FC layers'' network is the best performer. The result of this network on the complete dataset is mentioned at the $4^{th}$ row in Table~\ref{STL_accuracy}. Furthermore, the page level accuracy (evaluated on test dataset only) of ``Patch Level with multiple FC layers'' network by accumulating the patch level (results shown for word level also) posterior probabilities are mentioned in Table.~\ref{MTL_Full_Image}. It can be seen that thanks to our proposed voting scheme, mentioned in Section.~\ref{voting_scheme}, we have achieved very high page level accuracies in the case of ``Patch Level with multiple FC layers'' network.   
	 
	 	\begin{table}[h!]
	 	\small{
	 	\caption{Page level testing  accuracies of MTL based network using word and patch images }
	 	\centering
	 	\begin{tabular}{|P{2.1cm}|P{0.8cm} P{0.8cm} P{1.5cm} P{1.7cm}|}
	 		\hline
	 		&\textbf{Font Type} & \textbf{Font Size} & \textbf{Font Emphasis} & \textbf{Scanning Resolution} \\	
	 		\hline
	 		\textbf{Word Images} & \mycc 0.8879     &  \mycc 0.9129   &    \mycc 0.9392     &  \mycc 0.9935  \\
	 		\hline
	 	 	\textbf{Patch Images} & \mycc 0.9987     &  \mycc 1.00   &    \mycc 1.00     &  \mycc 1.00  \\
	 		\hline
	 	\end{tabular}
	 	\label{MTL_Full_Image}
	 	}
	 \end{table}
 	\vspace{-8mm}
	\section{Conclusions}
	\label{conclusion}
	\vspace{-2mm}
	In this work, we adhere to explore MTL to perform for $4$ document attribute classification tasks i.e. ``Font Type'', ``Font Size'', ``Font Emphasis'', ``Scanning Resolution'' recognition. In Table.~\ref{STL_accuracy}, we have shown that ``MTL'' based networks has outperformed ``STL'' based network. We further tried to combine the ``word and patch'' images together by proposing several MTL \& MI based network and weighted MTL \& MI based networks. But none of them performed better than ``MTL'' based network, using patch images as the input.   
	Hence, from all the above mentioned experiments, we can conclude that the proposed ``MTL'' based network, using patch images can attain high accuracy for these $4$ classification tasks.

%
	
	\bibliographystyle{model2-names}
	\bibliography{library.bib}

\begin{thebibliography}{18}
\expandafter\ifx\csname natexlab\endcsname\relax\def\natexlab#1{#1}\fi
\providecommand{\url}[1]{\texttt{#1}}
\providecommand{\href}[2]{#2}
\providecommand{\path}[1]{#1}
\providecommand{\DOIprefix}{doi:}
\providecommand{\ArXivprefix}{arXiv:}
\providecommand{\URLprefix}{URL: }
\providecommand{\Pubmedprefix}{pmid:}
\providecommand{\doi}[1]{\href{http://dx.doi.org/#1}{\path{#1}}}
\providecommand{\Pubmed}[1]{\href{pmid:#1}{\path{#1}}}
\providecommand{\bibinfo}[2]{#2}
\ifx\xfnm\relax \def\xfnm[#1]{\unskip,\space#1}\fi
\bibitem[{ocr()}]{ocr}
, .
\newblock \bibinfo{title}{{Tesseract OCR :}}.
\newblock \bibinfo{howpublished}{\url{https://github.com/tesseract-ocr/}}.
\bibitem[{dat()}]{datasetL3i}
, .
\newblock \bibinfo{title}{{TextCopies Dataset : }}.
\newblock
  \bibinfo{howpublished}{\url{http://navidomass.univ-lr.fr/TextCopies/}}.
\bibitem[{{Ben Moussa} et~al.(2010){Ben Moussa}, Zahour, Benabdelhafid and
  Alimi}]{BenMoussa2010}
\bibinfo{author}{{Ben Moussa}, S.}, \bibinfo{author}{Zahour, A.},
  \bibinfo{author}{Benabdelhafid, A.}, \bibinfo{author}{Alimi, A.M.},
  \bibinfo{year}{2010}.
\newblock \bibinfo{title}{{New features using fractal multi-dimensions for
  generalized Arabic font recognition}}.
\newblock \bibinfo{journal}{Pattern Recognition Letters} \bibinfo{volume}{31},
  \bibinfo{pages}{361--371}.
\bibitem[{Cloppet et~al.(2018)Cloppet, Eglin, Helias-Baron, Kieu, Vincent and
  Stutzmann}]{Cloppet2018}
\bibinfo{author}{Cloppet, F.}, \bibinfo{author}{Eglin, V.},
  \bibinfo{author}{Helias-Baron, M.}, \bibinfo{author}{Kieu, C.},
  \bibinfo{author}{Vincent, N.}, \bibinfo{author}{Stutzmann, D.},
  \bibinfo{year}{2018}.
\newblock \bibinfo{title}{{ICDAR2017 Competition on the Classification of
  Medieval Handwritings in Latin Script}}.
\newblock \bibinfo{journal}{Proceedings of the International Conference on
  Document Analysis and Recognition, ICDAR} \bibinfo{volume}{1},
  \bibinfo{pages}{1371--1376}.
\bibitem[{Das et~al.(2018)Das, Dantcheva and Bremond}]{das2018mitigating}
\bibinfo{author}{Das, A.}, \bibinfo{author}{Dantcheva, A.},
  \bibinfo{author}{Bremond, F.}, \bibinfo{year}{2018}.
\newblock \bibinfo{title}{Mitigating bias in gender, age and ethnicity
  classification: a multi-task convolution neural network approach}, in:
  \bibinfo{booktitle}{Proceedings of the European Conference on Computer Vision
  (ECCV)}, pp. \bibinfo{pages}{0--0}.
\bibitem[{Eskenazi et~al.(2015)Eskenazi, Gomez-Kr{\"{a}}mer and
  Ogier}]{Eskenazi2015}
\bibinfo{author}{Eskenazi, S.}, \bibinfo{author}{Gomez-Kr{\"{a}}mer, P.},
  \bibinfo{author}{Ogier, J.M.}, \bibinfo{year}{2015}.
\newblock \bibinfo{title}{{When document security brings new challenges to
  document analysis}}.
\newblock \bibinfo{journal}{Lecture Notes in Computer Science (including
  subseries Lecture Notes in Artificial Intelligence and Lecture Notes in
  Bioinformatics)} \bibinfo{volume}{8915}, \bibinfo{pages}{104--116}.
\bibitem[{Happy et~al.()Happy, Dantcheva, Das, Bremond, Zeghari and
  Robert}]{happy2020apathy}
\bibinfo{author}{Happy, S.}, \bibinfo{author}{Dantcheva, A.},
  \bibinfo{author}{Das, A.}, \bibinfo{author}{Bremond, F.},
  \bibinfo{author}{Zeghari, R.}, \bibinfo{author}{Robert, P.}, .
\newblock \bibinfo{title}{Apathy classification by exploiting task
  relatedness}, in: \bibinfo{booktitle}{2020 15th IEEE International Conference
  on Automatic Face and Gesture Recognition (FG 2020)(FG)}, pp.
  \bibinfo{pages}{733--738}.
\bibitem[{He et~al.(2016)He, Zhang, Ren and Sun}]{He2016}
\bibinfo{author}{He, K.}, \bibinfo{author}{Zhang, X.}, \bibinfo{author}{Ren,
  S.}, \bibinfo{author}{Sun, J.}, \bibinfo{year}{2016}.
\newblock \bibinfo{title}{{Deep residual learning for image recognition}}, in:
  \bibinfo{booktitle}{Proceedings of the IEEE Computer Society Conference on
  Computer Vision and Pattern Recognition}.
\newblock \href{http://arxiv.org/abs/1512.03385}{\tt arXiv:1512.03385}.
\bibitem[{Kang et~al.(2011)Kang, Grauman and Sha}]{kang2011learning}
\bibinfo{author}{Kang, Z.}, \bibinfo{author}{Grauman, K.},
  \bibinfo{author}{Sha, F.}, \bibinfo{year}{2011}.
\newblock \bibinfo{title}{Learning with whom to share in multi-task feature
  learning.}, in: \bibinfo{booktitle}{ICML}, pp. \bibinfo{pages}{521--528}.
\bibitem[{Krizhevsky et~al.(2017)Krizhevsky, Sutskever and
  Hinton}]{Krizhevsky2017}
\bibinfo{author}{Krizhevsky, A.}, \bibinfo{author}{Sutskever, I.},
  \bibinfo{author}{Hinton, G.E.}, \bibinfo{year}{2017}.
\newblock \bibinfo{title}{{ImageNet classification with deep convolutional
  neural networks}}, in: \bibinfo{booktitle}{NIPs}, pp.
  \bibinfo{pages}{84--90}.
\bibitem[{Lee et~al.(2016)Lee, Yang and Hwang}]{lee2016asymmetric}
\bibinfo{author}{Lee, G.}, \bibinfo{author}{Yang, E.}, \bibinfo{author}{Hwang,
  S.}, \bibinfo{year}{2016}.
\newblock \bibinfo{title}{Asymmetric multi-task learning based on task
  relatedness and loss}, in: \bibinfo{booktitle}{International Conference on
  Machine Learning}, pp. \bibinfo{pages}{230--238}.
\bibitem[{Long et~al.(2017)Long, Cao, Wang and Philip}]{long2017learning}
\bibinfo{author}{Long, M.}, \bibinfo{author}{Cao, Z.}, \bibinfo{author}{Wang,
  J.}, \bibinfo{author}{Philip, S.Y.}, \bibinfo{year}{2017}.
\newblock \bibinfo{title}{Learning multiple tasks with multilinear relationship
  networks}, in: \bibinfo{booktitle}{Advances in Neural Information Processing
  Systems}, pp. \bibinfo{pages}{1594--1603}.
\bibitem[{Pengcheng et~al.(2017)Pengcheng, Gang, Jiangqin and
  Baogang}]{Pengcheng2017}
\bibinfo{author}{Pengcheng, G.}, \bibinfo{author}{Gang, G.},
  \bibinfo{author}{Jiangqin, W.}, \bibinfo{author}{Baogang, W.},
  \bibinfo{year}{2017}.
\newblock \bibinfo{title}{{Chinese calligraphic style representation for
  recognition}}.
\newblock \bibinfo{journal}{International Journal on Document Analysis and
  Recognition} \bibinfo{volume}{20}, \bibinfo{pages}{59--68}.
\bibitem[{Shi et~al.(2016)Shi, Bai and Yao}]{Shi2016}
\bibinfo{author}{Shi, B.}, \bibinfo{author}{Bai, X.}, \bibinfo{author}{Yao,
  C.}, \bibinfo{year}{2016}.
\newblock \bibinfo{title}{{Script identification in the wild via discriminative
  convolutional neural network}}.
\newblock \bibinfo{journal}{Pattern Recognition} \bibinfo{volume}{52},
  \bibinfo{pages}{448--458}.
\bibitem[{Simard et~al.(2003)Simard, Steinkraus and Platt}]{Simard2003}
\bibinfo{author}{Simard, P.Y.}, \bibinfo{author}{Steinkraus, D.},
  \bibinfo{author}{Platt, J.C.}, \bibinfo{year}{2003}.
\newblock \bibinfo{title}{{Best practices for convolutional neural networks
  applied to visual document analysis}}.
\newblock \bibinfo{journal}{Proceedings of the International Conference on
  Document Analysis and Recognition, ICDAR} \bibinfo{volume}{2003-Janua},
  \bibinfo{pages}{958--963}.
\bibitem[{Tao et~al.(2016)Tao, Lin, Jin and Li}]{Tao2016}
\bibinfo{author}{Tao, D.}, \bibinfo{author}{Lin, X.}, \bibinfo{author}{Jin,
  L.}, \bibinfo{author}{Li, X.}, \bibinfo{year}{2016}.
\newblock \bibinfo{title}{{Principal Component 2-D Long Short-Term Memory for
  Font Recognition on Single Chinese Characters}}.
\newblock \bibinfo{journal}{IEEE Transactions on Cybernetics}
  \bibinfo{volume}{46}, \bibinfo{pages}{756--765}.
\bibitem[{Tensmeyer et~al.(2017)Tensmeyer, Saunders and
  Martinez}]{Tensmeyer2017}
\bibinfo{author}{Tensmeyer, C.}, \bibinfo{author}{Saunders, D.},
  \bibinfo{author}{Martinez, T.}, \bibinfo{year}{2017}.
\newblock \bibinfo{title}{{Convolutional Neural Networks for Font
  Classification}}.
\newblock \bibinfo{journal}{Proceedings of the International Conference on
  Document Analysis and Recognition, ICDAR} \bibinfo{volume}{1},
  \bibinfo{pages}{985--990}.
\newblock \href{http://arxiv.org/abs/1708.03669}{\tt arXiv:1708.03669}.
\bibitem[{Wu et~al.(2014)Wu, Jiang, Wang, Pu and Xue}]{wu2014exploring}
\bibinfo{author}{Wu, Z.}, \bibinfo{author}{Jiang, Y.G.}, \bibinfo{author}{Wang,
  J.}, \bibinfo{author}{Pu, J.}, \bibinfo{author}{Xue, X.},
  \bibinfo{year}{2014}.
\newblock \bibinfo{title}{Exploring inter-feature and inter-class relationships
  with deep neural networks for video classification}, in:
  \bibinfo{booktitle}{Proceedings of the 22nd ACM international conference on
  Multimedia}, \bibinfo{organization}{ACM}. pp. \bibinfo{pages}{167--176}.

\end{thebibliography}
	
	\large{{\color{blue}The paper is under consideration at "Pattern Recognition Letters" journal}}

	\pagebreak
	\renewcommand{\thetable}{S\arabic{table}}
	\setcounter{section}{0}
	\renewcommand{\thesection}{S\Roman{section}}
	
	\section*{ Supplementary Materials}
	\subsection{Implementation details of various proposed networks}
	\label{implement_details}
	In this section, we describe the implementation details of each of the networks which are used for the experiments. 
	We train the proposed network (mentioned in Section~\ref{proposed_method}) by using a pre-trained (trained on \emph{ImageNet} data-set) \emph{ResNet50} model. The model is trained by using stochastic gradient descent with momentum takes as $0.9$, multiplicative factor of learning rate decay (denoted by $\gamma$) is taken as $0.1$, step size is taken as $10$,  and initial learning rate is taken as $0.0001$. We use batches of size of $200$ and weight decay $0.0001$. The model is trained for $20000$ iterations per epoch and the loss function is cross-entropy with a uniform weighing scheme. We have used \textit{GeForce Titan RTX GPUs} with \textit{11GB} of RAM capacity per card. For faster execution, we have used $3$ GPU cards in parallel. 
	
	\subsubsection{Multi task network for segmented words or patches}
	For this case, almost same network parameters are maintained except the batch size is taken as $800$ for word images and $500$ for patch images (decided based on the memory capacity per GPU card). 
	
	\subsubsection{Multi task network for combined segmented words and patches}
	For this case, almost same network parameters are maintained except the batch size is taken as $100$ for both the \textit{multi task and multi instance learning} and as well as for \textit{weighted multi task and multi instance learning}. 
	
	\subsection{Training and Validation accuracies of MTL networks}
	\label{train_val_accura}
	In the following Fig.\ref{fig:multitask_patch_word}, we have shown the training and validation accuracies of MTL based word level and patch level networks. It can be visible that the accuracies of the $4$ tasks increases with the iteration of each epochs. 
	\begin{figure}[!htb]
		\centering
		\begin{minipage}{\linewidth}
			\centering{\includegraphics[trim = 4.2cm 1.9cm 3.4cm 1.3cm, clip,  scale = 0.21]{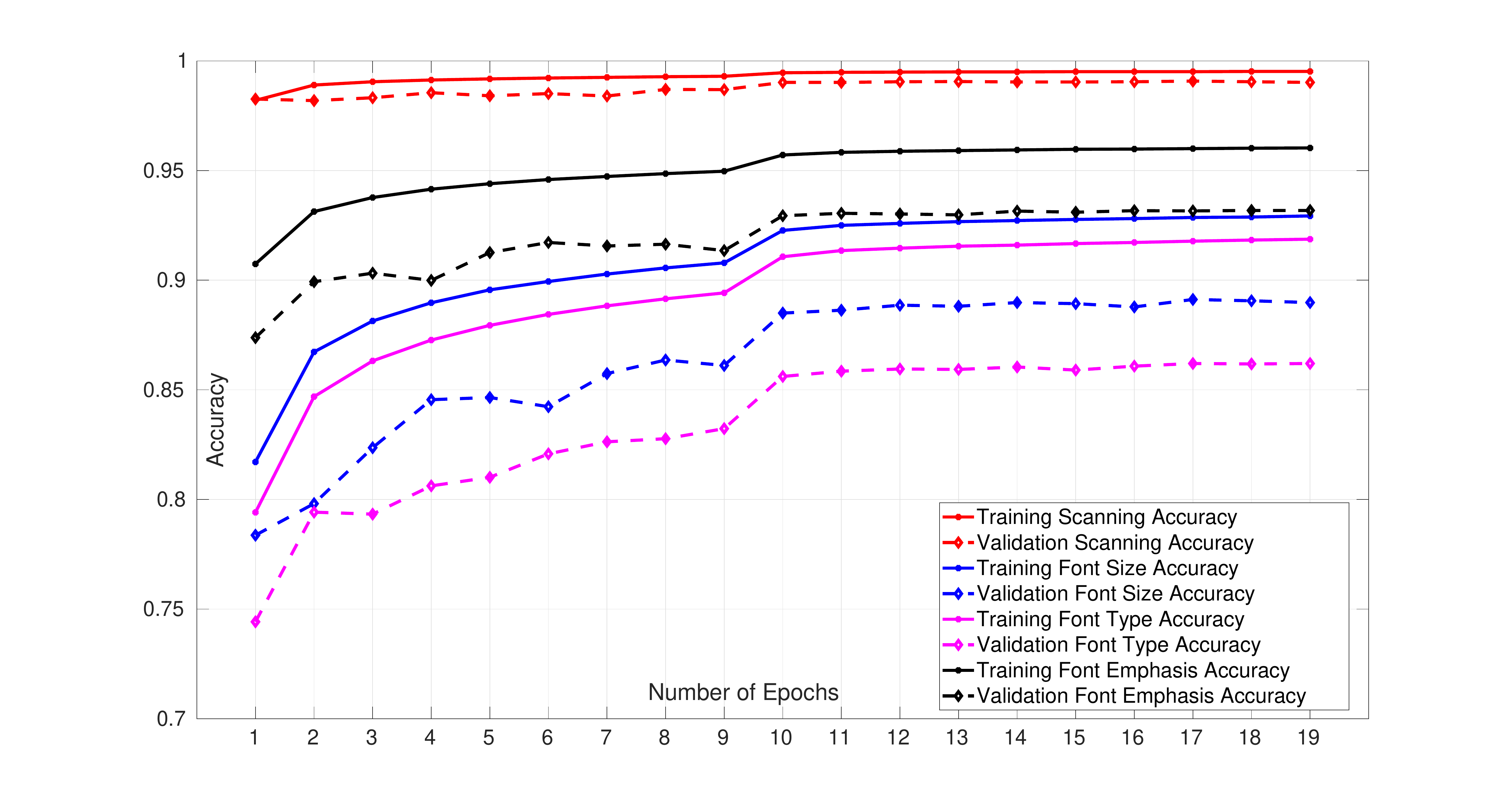}}
			\\{\hspace{5mm}\small{(a)}}
		\end{minipage}\\
		\begin{minipage}{\linewidth}
			\centering{\includegraphics[trim = 4.2cm 1.6cm 3.4cm 1.3cm, clip,  scale = 0.21]{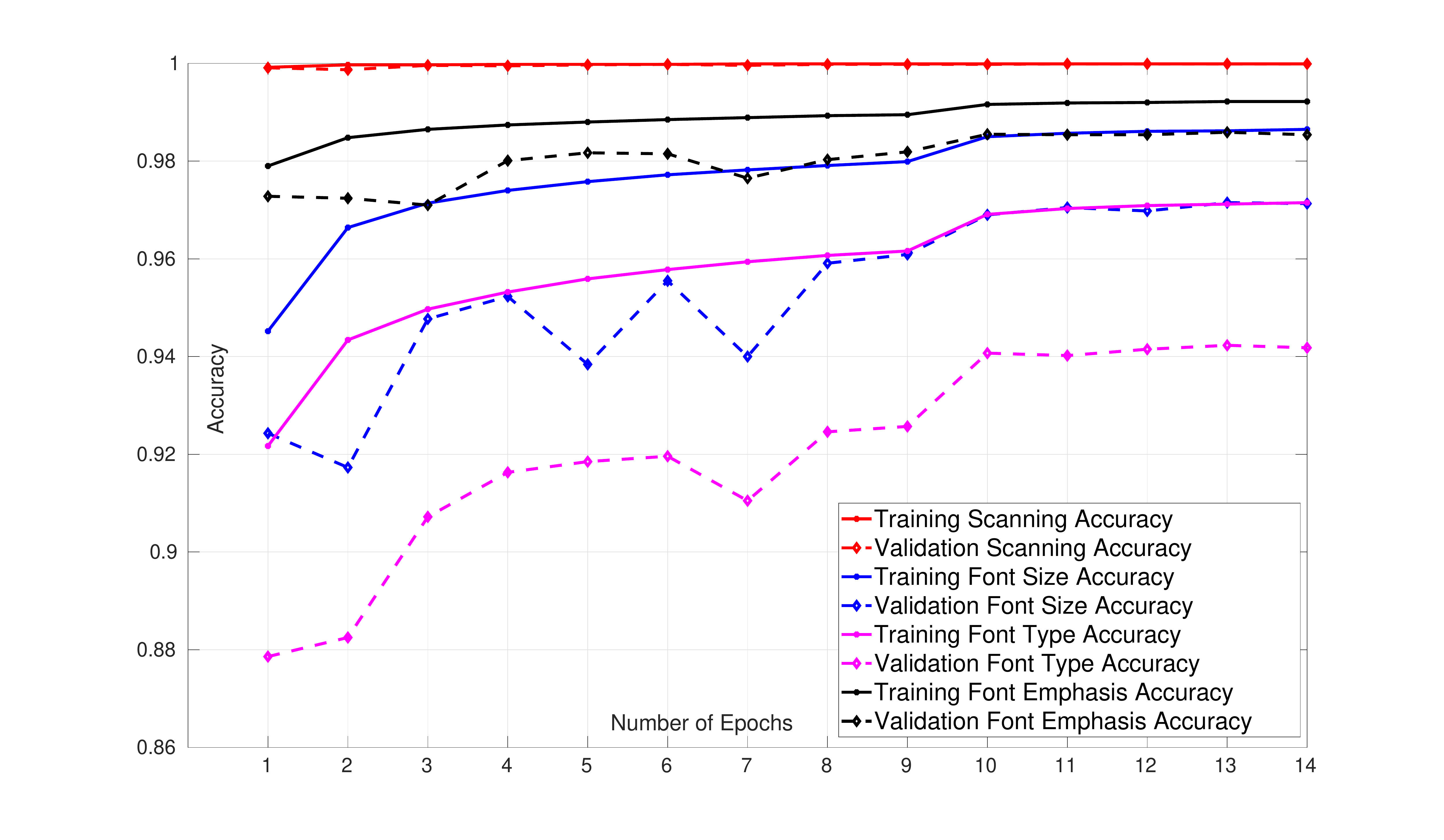}}
			\\{\vspace{-2mm}\hspace{16mm}\small{(b)}}
		\end{minipage}
		\caption{Training and validation accuracies of MTL based network for ``Font type'', ``Font size'', ``Font emphasis'' and ``Scanning resolution'' tasks : (a) Taking word images as input  (b) Taking patch images as input.
		}
		\label{fig:multitask_patch_word}
	\end{figure}

	\subsection{Further experiments on MTL based networks}
	\label{further_MTL}
	In continuation with the experiments, mentioned in Table.~\ref{STL_accuracy}, we further has performed some more experiments. In another similar kind of network, we adopt the ``VggNet'' \citep{He2016} like architecture instead of ``AlexNet'' like architecture (the architectural details remain same as the one, mentioned in Section~\ref{alexnet_MTL_MI}). In this network, we have treated the $2048$ pre-trained features as $1D$ channel. Then this $1D$ features are passed through the following $2$ blocks of: 
	\fbox{\parbox{\dimexpr\linewidth-2\fboxsep-2\fboxrule\relax}{\centering 
			\strut \textbf{CONV}($in_{ch}^1 \rightarrow out_{ch}^1; k^1$)-\textbf{RELU}-\textbf{CONV} ($in_{ch}^2 \rightarrow out_{ch}^2; k^2$)-\textbf{RELU}-\textbf{MaxPool}(n)
	}} 
	The parameters are taken as follows in these $2$ consequent blocks i.e. for:
	
	\textbf{$1^{st}$ block:} 
	
	$in_{ch}^1 = 2$, $out_{ch}^1=64$, $k^1=3$; 
	
	$in_{ch}^2 = 64$, $out_{ch}^2=64$, $k^2=3$. 
	
	\vspace{2mm}
	\textbf{$2^{nd}$ block:} 
	
	$in_{ch}^1 = 64$, $out_{ch}^1=128$, $k^1=3$ 
	
	$in_{ch}^2 = 128$, $out_{ch}^2=128$, $k^2=3$.
	
	\vspace{2mm}
	Then the output of this $2^{nd}$ block is passed through the following $3$ blocks of:
	
	\fbox{\parbox{\dimexpr\linewidth-0\fboxsep-2\fboxrule\relax}{\centering 
			\strut \textbf{CONV}($in_{ch}^1 \rightarrow out_{ch}^1; k^1$)-\textbf{RELU}-\textbf{CONV} ($in_{ch}^2 \rightarrow out_{ch}^2; k^2$)-\textbf{RELU}-\textbf{CONV} ($in_{ch}^3 \rightarrow out_{ch}^3; k^3$)-\textbf{RELU}-\textbf{CONV} ($in_{ch}^4 \rightarrow out_{ch}^4; k^4$)-\textbf{RELU}-\textbf{MaxPool}($\mathcal{K}$)
	}}
	The parameters are taken as follows in these $3$ consequent blocks i.e. 
	
	\textbf{for $3^{rd}$ block}: 
	
	$in_{ch}^1 = 128$, $out_{ch}^1=256$, $k^1=3$; 
	
	$in_{ch}^2 = 256$, $out_{ch}^2=256$, $k^2=3$;
	
	$in_{ch}^3 = 256$, $out_{ch}^3=256$, $k^3=3$; 
	
	$in_{ch}^4 = 256$, $out_{ch}^4=256$, $k^4=3$; $\mathcal{K}=2$. 
	
	\vspace{2mm}
	For the \textbf{$4^{th}$ block}: 
	
	$in_{ch}^1 = 256$, $out_{ch}^1=512$, $k^1=3$; 
	
	$in_{ch}^2 = 512$, $out_{ch}^2=512$, $k^2=3$;
	 
	$in_{ch}^3 = 512$, $out_{ch}^3=512$, $k^3=3$; 
	
	$in_{ch}^4 = 512$, $out_{ch}^4=512$, $k^4=3$; $\mathcal{K}=2$. 
	
	\vspace{2mm}
	For the \textbf{$5^{th}$ block}: 
	
	$in_{ch}^1 = 512$, $out_{ch}^1=512$, $k^1=3$; 
	
	$in_{ch}^2 = 512$, $out_{ch}^2=512$, $k^2=3$; 
	
	$in_{ch}^3 = 512$, $out_{ch}^3=512$, $k^3=3$; 
	
	$in_{ch}^4 = 512$, $out_{ch}^4=512$, $k^4=3$; $\mathcal{K}=2$. 
	
	\begin{table}[h!]
		\small{
		\caption{Training, validation and testing accuracies of MTL based network for word and patch level images }
		\centering
		\begin{tabular}{|P{2cm}|P{0.8cm} P{0.8cm} P{1.5cm} P{1.7cm}|}
			\hline
			\multicolumn{5}{|c|}{\textbf{Top-1 accuracy of MTL}} \\ \hline
			& \textbf{Font Type} & \textbf{Font Size} & \textbf{Font Emphasis} & \textbf{Scanning Resolution} \\ \hline
			
			\multirow{3}{*}{\shortstack{\textbf{Word} Level \\ with \textbf{VggNet} \\ layers}} & \mycc 0.1762     &  \mycc 0.3485    &    \mycc 0.2522     &  \mycc 0.3459  \\
			& \mybb 0.1740     & \mybb 0.3482    &   \mybb 0.2515     & \mybb 0.3479 \\
			& \myrr 0.1751    & \myrr  0.3508  &   \myrr 0.2484    & \myrr 0.3507 \\
			\hline

			\multicolumn{5}{ c }{\textbf{   }} \\ \hline

			\multirow{3}{*}{\shortstack{\textbf{Patch} Level \\ with \textbf{VggNet} \\ layers}} & \mycc 0.1671     &  \mycc 0.3314    &    \mycc 0.2519     &  \mycc 0.3327  \\
			& \mybb 0.1660     & \mybb 0.3319    &   \mybb 0.2488     & \mybb 0.3319 \\
			& \myrr 0.1677    & \myrr 0.3003   &   \myrr 0.2446    & \myrr  0.3693\\
			\hline
			
		\end{tabular}
		\label{supp_MTL_accuracy_1}
		}
	\end{table}	
	
	The output of the $5^{th}$ block is a $2D$ features of size $512 \times 64$ which is passed through the ``Average Pooling'' layer\footnote{here we have used ``Adaptive Average Pooling'' algorithm from PyTorch library. For more details, see : \url{https://pytorch.org/cppdocs/api/classtorch_1_1nn_1_1_adaptive_avg_pool1d.html} }, having a kernel of size $6$. This makes the $2D$ features to get flattened and reduced into the $1D$ feature of size $3072$. Then $3072$ features are passed through following layers:
	\fbox{\parbox{\dimexpr\linewidth-0\fboxsep-2\fboxrule\relax}{\centering 
			\strut \textbf{DropOut}-\textbf{FC}($(512\times6) \rightarrow 768$)-\textbf{RELU}-\textbf{DropOut}-\textbf{FC}($768 \rightarrow 384$)-\textbf{RELU}-\textbf{FC}($384 \rightarrow 192$)
	}}   
	After passing through the above mentioned layers, we obtain $192$ number of features which are finally connected to $4$ output heads, corresponding to $4$ individual tasks.  We have named this network as ``Early concat VggNet like conv. layers'' and it's results are shown in Table~\ref{supp_MTL_accuracy_1}.   
	
	\subsubsection{Results and Discussion}
	The results of above described network mentioned in Table.~\ref{supp_MTL_accuracy_1}, where either word images (named as ``Word Level with VggNet layers'') or patch images (named as ``Patch Level with VggNet layers'') are taken as input to the network. It can be visible from these results that the frequently arising model over fitting problem can be clearly resolved in this network but the accuracy of this network for both types of input i.e. either word images or patch images as input, shows inferior accuracies compared to the other networks, shown in Table.~\ref{MTL_accuracy_1}.

	\subsection{Further experiments on MTL \& MI based networks}
	\label{supp_MTL_MI}
	In continuation with the experiments, mentioned in Section~\ref{accu_multi_task_multi_instance}, where we have tried to replace the late concatenation of patch and word level features by the early concatenation of these features. In the Section~\ref{accu_multi_task_multi_instance}, we have discussed such a network where the early concatenated $4096$ features are passed through several liner layers, before getting connected to $4$ output heads. The detail of this network is mentioned below in Section~\ref{early_concat_MTL_MI}.
	
	\subsubsection{Early concat multiple FC layers network}
	\label{early_concat_MTL_MI}
	In this network, $2048$ number of pre-trained ``ResNet-50'' features of word and patch images are concatenated to generate $4096$ features,
	which are sequentially passed through several layers like : 
	\fbox{ \strut \textbf{FC} ($4096 \rightarrow 2048$)-\textbf{RELU}-\textbf{BN}}; \fbox{ \strut \textbf{FC} ($2048 \rightarrow 1024$)-\textbf{RELU}-\textbf{BN}};~\fbox{ \strut \textbf{FC} ($1024 \rightarrow 512$)-\textbf{RELU}-\textbf{BN}}; \fbox{ \strut \textbf{FC} ($512 \rightarrow 256$)-\textbf{RELU}-\textbf{BN}};~\fbox{ \strut \textbf{FC} ($256 \rightarrow 128$)-\textbf{RELU}-\textbf{BN}}; ~\fbox{ \strut \textbf{FC} ($128 \rightarrow 64$)-\textbf{RELU}-\textbf{BN}}; \fbox{ \strut \textbf{FC} ($64 \rightarrow 32$)-\textbf{RELU}-\textbf{BN}}; \fbox{ \strut \textbf{FC} ($32 \rightarrow 16$)-\textbf{RELU}-\textbf{BN}} and finally get connected to $4$ output heads. The results of this network are shown at the $2^{nd}$ row of Table~\ref{MTL_accuracy_2} and are named as ``Early concat multiple FC layers'' network. 
	
	Although, by using this network, the validation and testing accuracies got highly improved (see Table.~\ref{MTL_accuracy_2}) but still there remains a strong gap between the training and validation/testing accuracies. We suspected that may be the pipeline connection of several FC layers could be the possible reason of this gap between training and validation/testing accuracies of $4$ tasks, hence we tried to reduce the number of FC layers.  
	
	\subsubsection{Reduced number of FC layers for MTL \& MI based network}
	\label{reduced_LL_MTL_MI}
	Consequently, such a network is explained here. In this network, the pre-trained concatenated $4096$ features are sequentially passed through smaller number of FC layers such as: \fbox{ \strut \textbf{FC} ($4096 \rightarrow 2048$)-\textbf{RELU}-\textbf{BN}} which is then directly connected to $4$ output heads. This network is named as ``Early concat less FC layers\_1''.
	
	We further modified the network by sequentially passing through the concatenated $4096$ features by $2$ FC layers as : \fbox{ \strut \textbf{FC} ($4096 \rightarrow 2048$)-\textbf{RELU}-\textbf{BN}}; \fbox{ \strut \textbf{FC} ($2048 \rightarrow 1024$)-\textbf{RELU}-\textbf{BN}}; which then directly connected to $4$ output heads and we named this network as ``Early concat less FC layers\_2''. The results of these two networks are mentioned in $1^{st}$ and $2^{nd}$ rows of Table~\ref{supp_MTL_accuracy_2} respectively. It can be seen from the validation and testing results of these two networks that the results doesn't get improved for all the $4$ tasks and the problem of ``model over fitting'' still remains.
	
	\begin{table}[h!]
		\small{
		\caption{Training, validation and testing accuracies of MTL and MI based network by combining word and patch level images }
		\centering
		\begin{tabular}{|P{2cm}|P{0.8cm} P{0.8cm} P{1.5cm} P{1.7cm}|}
			\hline
			\multicolumn{5}{c}{\shortstack{\textbf{Top-1 accuracy of MTL by combining} \\ \textbf{word and patch images}}} \T \\ \hline
			& \textbf{Font Type} & \textbf{Font Size} & \textbf{Font Emphasis} & \textbf{Scanning Resolution} \\ \hline

			\multirow{3}{*}{\shortstack{\textbf{Early} concat \\ \textbf{less} FC\\ layers\_1}} & \mycc 0.9669     &  \mycc 0.9642    &    \mycc  0.9817    &  \mycc  0.9977  \\
			& \mybb 0.4865     & \mybb 0.5204    &   \mybb 0.5644     & \mybb 0.9692 \\
			& \myrr 0.5049     & \myrr 0.4258    &   \myrr 0.5631     & \myrr 0.8041 \\
			\hline
			\multirow{3}{*}{\shortstack{\textbf{Early} concat \\ \textbf{less} FC\\ layers\_2}} & \mycc 0.9715     &  \mycc 0.9714    &    \mycc 0.9836     &  \mycc 0.9976  \\
			& \mybb 0.4976     & \mybb 0.4690    &   \mybb 0.5661     & \mybb 0.9739 \\
			& \myrr 0.4975    & \myrr  0.4052   &   \myrr 0.6156     & \myrr 0.7831 \\
			\hline
			\hline
			
			\multirow{3}{*}{\shortstack{\textbf{Early} concat \\ \textbf{VggNet} like \\ Conv. layers}} & \mycc 0.1708     &  \mycc 0.3457    &    \mycc 0.2486     &  \mycc 0.3448  \\
			& \mybb 0.1723     & \mybb 0.3451    &   \mybb 0.2497     & \mybb 0.3452 \\
			& \myrr 0.1705    & \myrr 0.3683   &   \myrr 0.2450    & \myrr 0.3763  \\
			\hline
			\hline
		\end{tabular}
		\label{supp_MTL_accuracy_2}
		}
	\end{table}
	
	\subsubsection{AlexNet like convolution layers for MTL \& MI based network}
	\label{alexnet_MTL_MI}
	To handle the problem of ``model over fitting'', in Section~~\ref{accu_multi_task_multi_instance}, we proposed to replace the FC layers by convolutional layers, e.g. adapting ``AlexNet'' like architecture. 	We have treated the $2048$ pre-trained features of patch and $2048$ pre-trained features of word images as two channels of $1D$ vector. Then these two channels are given as input and are passed through the following $5$ blocks of:\fbox{ \strut \textbf{CONV} ($in_{ch} \rightarrow out_{ch}; k$)-\textbf{BN}-\textbf{RELU}}; where the parameters i.e. $in_{ch}$ represents number of input channels, $out_{ch}$ represents number of output channels and $k$ represents the kernel size. The parameters are taken as follows in these $5$ consequent blocks i.e. for:
	
	\textbf{$1^{st}$ block:} $in_{ch} = 2$, $out_{ch}=64$, $k=11$; 
	
	\textbf{$2^{nd}$ block:} $in_{ch} = 64$, $out_{ch}=192$, $k=5$; 
	
	\textbf{$3^{rd}$ block:} $in_{ch} = 192$, $out_{ch}=384$, $k=3$; 
	
	\textbf{$4^{th}$ block:} $in_{ch} = 384$, $out_{ch}=256$, $k=3$; 
	
	\textbf{$5^{th}$ block:} $in_{ch} = 256$, $out_{ch}=256$, $k=3$; 
	
	Finally, the output of $5^{th}$ block ($192$ features) are finally get connected to $4$ output heads.
		
	\subsubsection{VggNet like convolution layers for MTL \& MI based network}
	\label{vggnet_MTL_MI}
	
	In another similar kind of network, we adopt the ``VggNet'' \citep{He2016} like architecture instead of ``AlexNet'' like architecture. In this network also, we have treated the $2048$ pre-trained features of patch and $2048$ pre-trained features of word images as two $1D$ channels. Then these two channels of features are passed through the following $2$ blocks of: 
	\fbox{\parbox{\dimexpr\linewidth-2\fboxsep-2\fboxrule\relax}{\centering 
			\strut \textbf{CONV}($in_{ch}^1 \rightarrow out_{ch}^1; k^1$)-\textbf{RELU}-\textbf{CONV} ($in_{ch}^2 \rightarrow out_{ch}^2; k^2$)-\textbf{RELU}-\textbf{MaxPool}(n)
	}} 
	Then the output of this $2^{nd}$ block is passed through the following $3$ blocks of:
	
	\fbox{\parbox{\dimexpr\linewidth-0\fboxsep-2\fboxrule\relax}{\centering 
			\strut \textbf{CONV}($in_{ch}^1 \rightarrow out_{ch}^1; k^1$)-\textbf{RELU}-\textbf{CONV} ($in_{ch}^2 \rightarrow out_{ch}^2; k^2$)-\textbf{RELU}-\textbf{CONV} ($in_{ch}^3 \rightarrow out_{ch}^3; k^3$)-\textbf{RELU}-\textbf{CONV} ($in_{ch}^4 \rightarrow out_{ch}^4; k^4$)-\textbf{RELU}-\textbf{MaxPool}($\mathcal{K}$)
	}}
		 
	\begin{table}[h!]
		\small{
		\caption{Training, validation and testing  accuracies of MTL based network by combining normal patch and noisy patch level images }
		\centering
		\begin{tabular}{|P{2cm}|P{0.8cm} P{0.8cm} P{1.5cm} P{1.7cm}|}
			\hline
			\multicolumn{5}{c}{\shortstack{\textbf{Top-1 accuracy of MTL by combining} \\ \textbf{ normal  patch and noisy patch images }}} \T \\ \hline
			& \textbf{Font Type} & \textbf{Font Size} & \textbf{Font Emphasis} & \textbf{Scanning Resolution} \\ \hline	
			\hline
			\multirow{3}{*}{ \shortstack{\textbf{Early} concat \\ \textbf{AlexNet} like \\ Conv. layers} } & \mycc 0.2995     &  \mycc 0.5348    &    \mycc 0.5389     &  \mycc 0.9667  \\
			& \mybb 0.2883     & \mybb 0.5312    &   \mybb 0.6386     & \mybb 0.9914 \\
			& \myrr 0.2592    & \myrr 0.5067   &   \myrr  0.6422   & \myrr 0.9691  \\
			\hline
			\multirow{3}{*}{ \shortstack{\textbf{Early} concat \\ \textbf{VggNet} like \\ Conv. layers} } & \mycc 0.1740     &  \mycc 0.3457    &    \mycc 0.2521     &  \mycc 0.3452  \\
			& \mybb 0.1748     & \mybb 0.3495    &   \mybb 0.2524     & \mybb 0.3495 \\
			& \myrr 0.1713    & \myrr 0.3695   &   \myrr 0.2435    & \myrr 0.3762 \\
			\hline
		\end{tabular}
		\label{MTL_accuracy_noisy_patch}
		}
	\end{table}

	The parameters of these $5$ blocks remain same as it is mentioned in Section~\ref{further_MTL}. In the same fashion, the output of the $5^{th}$ block is a $2D$ features of $512 \times 64$ dimension which is passed through the ``Average Pooling'' layer, having a kernel of size $6$. This makes the $2D$ features to get flattened and reduced into the $1D$ feature of size $3072$. Then $3072$ features are passed through following FC layers:
	\fbox{\parbox{\dimexpr\linewidth-0\fboxsep-2\fboxrule\relax}{\centering 
			\strut \textbf{DropOut}-\textbf{FC}($(512\times6) \rightarrow 768$)-\textbf{RELU}-\textbf{DropOut}-\textbf{FC}($768 \rightarrow 384$)-\textbf{RELU}-\textbf{FC}($384 \rightarrow 192$)
	}}   
	After passing through the above mentioned layers, we obtain $192$ number of features which are finally connected to $4$ output heads, corresponding to $4$ individual tasks.  We have named this network as ``Early concat VggNet like conv. layers'' and the results are shown in $3^{rd}$ row of Table~\ref{supp_MTL_accuracy_2}. 
	
	It can be visible from results that the annoying problem of ``model over fitting'' is also resolved by this network but the overall accuracies of all the $4$ tasks are inferior to it's counterpart i.e. ``Early concat AlexNet like Conv. layers'' network (see Table.~\ref{MTL_accuracy_2}). 

	\subsubsection{Using noisy patch as input in MTL \& MI based network}
	As it is visible in Table.~\ref{MTL_accuracy_2} and are mentioned in Section~\ref{accu_multi_task_multi_instance} that we get highly annoyed by ``model over fitting'' problem when we tried to combine word and patch images together as the inputs in proposed three networks i.e. ``Late concat multiple FC layers'', ``Early concat multiple FC layers'', ``Early concat AlexNet like Conv. layers''. We have tried several tricks and strategies to overcome the problem of ``model over fitting'', which are mentioned in Section~\ref{accu_multi_task_multi_instance} and Section~\ref{reduced_LL_MTL_MI} and \ref{vggnet_MTL_MI}. But none of these proposed network architecture could outperform the results of MTL based networks, which uses either word images (``Word Level with multiple FC layers'' network) or patch images (``Patch Level with multiple FC layers'' network) shown in Table.~\ref{MTL_accuracy_1}. It can also be seen from Table.~\ref{MTL_accuracy_1} that ``Patch Level with multiple FC layers'' network has shown superior accuracy and has overcome the ``model over fitting'' problem, compared to ``Word Level with multiple FC layers'' network. Which inherently imply that word images as the input may be the possible offender for this ``model over fitting'' problem as well as the reason of decline in accuracy of the results, shown in Table.~\ref{MTL_accuracy_2} and Table.\ref{supp_MTL_accuracy_2}.
	
	Hence, in the following setup, we have replaced cropped word images by noisy patch images (get inspired from ``denoising auto encoder'' \footnote{ I. Goodfellow, Y. Bengio, A. Courville, Deep Learning (2016), The MIT Press}). Hence, we tested the performance of ``Early concat AlexNet like Conv. layers'' network (described in $2^{nd}$ last paragraph in Section~\ref{accu_multi_task_multi_instance}) and ``Early concat VggNet like Conv. layers'' network (described in Section~\ref{vggnet_MTL_MI}) by considering patch images as the $1^{st}$ input and noisy version of the same patch images (we have applied standard Gaussian noise\footnote{\url{https://gist.github.com/Prasad9/28f6a2df8e8d463c6ddd040f4f6a028a\#gistcomment-2857098}} of mean=$0$ and standard deviation = $\sqrt 0.1$) as the $2^{nd}$ input of the network. The data is partially corrupted by noises and are added to the input vector in a stochastic manner. Then the network is trained to correctly classify even if it is trained with noisy images.    
	
	The results of these networks are shown in Table.\ref{MTL_accuracy_noisy_patch}. It can be visible from the results that the problem of ``model over fitting'' is resolved here also. Moreover, the results of ``Early concat AlexNet like Conv. layers'' network got highly improved in comparison with it's counterpart in Table.~\ref{MTL_accuracy_2}. Whereas, the results of ``Early concat VggNet like Conv. layers'' network from Table.\ref{MTL_accuracy_noisy_patch} remains same as it's counter part, mentioned in Table.~\ref{supp_MTL_accuracy_2}. Most probable reason could be the inherent architecture of ``VggNet'', which is able to overcome the ``model over fitting'' problem, irrespective of word images (which is the most probable offender for ``model over fitting'' problem) as input to the network. But in particular, both of these networks couldn't outperform the ``Patch Level with multiple FC layers'' network, mentioend in Table.~\ref{MTL_accuracy_1}.    

	\begin{table}[h!]
		\small{
		\caption{Training, validation and testing  accuracies of weighted MTL based network by combining normal patch and noisy patch level images }
		\centering
			\begin{tabular}{|P{2.2cm}|P{0.8cm} P{0.8cm} P{1.5cm} P{1.7cm}|}
			\hline
			\multicolumn{5}{c}{\shortstack{\textbf{Top-1 accuracy of weighted MTL by combining} \\ \textbf{ word and patch level images }}} \T \\ \hline
			& \textbf{Font Type} & \textbf{Font Size} & \textbf{Font Emphasis} & \textbf{Scanning Resolution} \\ \hline	
			\hline
			\multirow{3}{*}{ \shortstack{\textbf{Late} concat \\ \textbf{multiple} FC \\ layers} } & \mycc 0.5025     &  \mycc 0.7604    &    \mycc 0.8432     &  \mycc 0.9905  \\
			& \mybb 0.4469     & \mybb 0.6948    &   \mybb 0.8364     & \mybb 0.9884 \\
			& \myrr 0.3017    & \myrr 0.4366   &   \myrr 0.5966     & \myrr 0.7374  \\
			\hline
			\multirow{3}{*}{ \shortstack{\textbf{Late} concat \\ \textbf{AlexNet} like \\ Conv. layers} } & \mycc 0.2946     &  \mycc 0.5657    &    \mycc 0.5791     &  \mycc 0.9823  \\
			& \mybb 0.2614     & \mybb 0.5224    &   \mybb 0.6412     & \mybb 0.9892 \\
			& \myrr 0.2012   & \myrr 0.4742   &   \myrr  0.4517   & \myrr 0.9302 \\
			\hline
		\end{tabular}
		\label{MTL_weighted_noisy_patch}
		}
	\end{table}
	
	\section{Further experiments on weighted MTL \& MI based learning}
	In the following section, we have explained further details about several experiments regarding weighted MTL \& MI based learning. 
	\subsection{AlexNet like Weighted Multi-task and Multi-instance Learning}
	\label{alex_net_weighted_multi_task_multi_instance}
	Here we have mentioned the details about ``Late concat AlexNet like Conv. layers'' network, whose results are given in Table.~\ref{MTL_weightedaccuracy}. 
	After obtaining $2048$ pretrained ``ResNet-50'' features from word and patch images, we pass them through a FC layer of: \fbox{ \strut \textbf{FC} ($2048 \rightarrow 512$)} to reduce the dimension of feature. After obtaining $512$ features of word image network and $512$ features of patch image network, we treat each of them as a channel of $1D$ vector. Then each of these channel are passed through the $5$ blocks of: \fbox{ \strut \textbf{CONV} ($in_{ch} \rightarrow out_{ch}; k$)-\textbf{BN}-\textbf{RELU}}; where the parameters i.e. $in_{ch}$ represents number of input channels, $out_{ch}$ represents number of output channels and $k$ represents the kernel size. The parameters are taken as before like ``Late concat AlexNet like conv. layers'' network, mentioned in Section~\ref{accu_multi_task_multi_instance}. 
	
	After passing through the $5^{th}$ block, we obtain the $2D$ features of $256\times124$ dimension. This $2D$ feature is passed through the ``Average Pooling'' layer (having kernel of size $6$), which scale down the feature into of dimension $256 \times 6$. Then this reduced dimensional feature is flatten\footnote{we have used ``nn.Flatten()'' function of PyTorch library to flatten the feature. For more details, please see : \url{https://pytorch.org/docs/stable/generated/torch.nn.Flatten.html}} to get $1D$ features of size $1536$ which is then passed through 3 blocks of following layers: \fbox{ \strut \textbf{DropOut}($\mathcal{P}=50$)-\textbf{FC} ($1536 \rightarrow 768$)-\textbf{RELU}}; \fbox{ \strut \textbf{DropOut}($\mathcal{P}=50$)-\textbf{FC} ($768 \rightarrow 384$)-\textbf{RELU}}; \fbox{ \strut \textbf{FC} ($384 \rightarrow 192$)}; After passing through these $3$ blocks, we finally obtain $192$ features from the word network (using word images only) as well as from the patch network (using patch image only). The $192$ output features of word network get connected to $4$ output heads (for reference, see ``Word Multi-tasking block'' at the left of Fig.~\ref{fig:archi_resnet_combine_weighted}). Another set of $192$ features from patch images are also get connected to $4$ output heads (for reference, see ``Patch Multi-tasking block'' at the right of Fig.~\ref{fig:archi_resnet_combine_weighted}).  
	The weights are automatically computed in the same manner by initially combining $192$ features of ``word'' network and $192$ features of ``patch'' network together. Then these $384$ number of combined features are  individually connected to $4$ heads of: \fbox{ \strut \textbf{FC} ($384 \rightarrow 192$)-\textbf{RELU}}; \fbox{ \strut \textbf{FC} ($192 \rightarrow 96$)-\textbf{RELU}}; \fbox{ \strut \textbf{FC} ($96 \rightarrow 48$)-\textbf{RELU}}; \fbox{ \strut \textbf{FC} ($48 \rightarrow 24$)-\textbf{RELU}}; \fbox{ \strut \textbf{FC} ($24 \rightarrow 2$)-\textbf{RELU}} (take reference from the Fig.~\ref{fig:archi_resnet_combine_weighted} where you can see that each output head is outputting $2$ weight values). Hence, from each output head, we can get two weight values where the $1^{st}$ weight value is dedicated for the output of ``word multi-tasking block'' and the $2^{nd}$ weight value is dedicated for ``patch multi-tasking block''. 
	After obtaining two weight values from each output head, dedicated to \emph{Font Emphasis}, \emph{Font Type}, \emph{Font Size} and \emph{Scanning Resolution} tasks, these ones are multiplied  and averaged with the outputs of each output heads of ``word multi-tasking block'' as well as ``patch multi-tasking block'' in the same manner as it is shown and described in Fig.~\ref{fig:archi_resnet_combine_weighted}. The results of this network is mentioned in Table.~\ref{MTL_weightedaccuracy} and the corresponding description is given in Section~\ref{weighted_accu_multi_task_multi_instance}.
	
	In the following Table.~\ref{MTL_weighted_no_softmax}, we have also experimented the effect of ``soft-max'' layer while computing the weight values from each output heads. Which means, we have removed the ``soft-max'' layers in both the network i.e. ``Late concat multiple FC layers'' and ``Late concat AlexNet like Conv. layers'' (see Table.~\ref{MTL_weightedaccuracy}) networks  from all $4$ output heads of ``Word Multi-tasking Block'' and ``Patch Multi-tasking Block'', shown in Fig.\ref{fig:archi_resnet_combine_weighted}. It can be clearly visible from the results in Table.~\ref{MTL_weighted_no_softmax} in comparison with the results, shown in Table.~\ref{MTL_weightedaccuracy} that when used during the weight computation, the ``soft-max'' layers  plays a vital role in improving accuracy.  
	\begin{table}[h!]
		\small{
		\caption{Training, validation and testing  accuracies of weighted MTL based network by combining word and patch level images (without soft-max layer for weight calculation) }
		\centering
		\begin{tabular}{|P{2.2cm}|P{0.8cm} P{0.8cm} P{1.5cm} P{1.7cm}|}
			\hline
			\multicolumn{5}{c}{\shortstack{\textbf{Top-1 accuracy of weighted MTL by combining} \\ \textbf{ word and patch level images }}} \T \\ \hline
			& \textbf{Font Type} & \textbf{Font Size} & \textbf{Font Emphasis} & \textbf{Scanning Resolution} \\ \hline	
			\hline
			\multirow{3}{*}{ \shortstack{\textbf{Late} concat \\ \textbf{multiple} FC \\ layers} } & \mycc 0.4887     &  \mycc 0.6970    &    \mycc 0.6511     &  \mycc 0.9224  \\
			& \mybb 0.2558     & \mybb 0.4053    &   \mybb 0.4091     & \mybb 0.7032 \\
			& \myrr 0.2417    & \myrr 0.3733   &   \myrr 0.3789     & \myrr 0.5368   \\
			\hline
			\multirow{3}{*}{ \shortstack{\textbf{Late} concat \\ \textbf{AlexNet} like \\ Conv. layers} } & \mycc 0.1824     &  \mycc 0.3823    &    \mycc 0.2631     &  \mycc 0.6090  \\
			& \mybb 0.1790     & \mybb 0.3701    &   \mybb 0.2476     & \mybb 0.5713 \\
			& \myrr 0.1828   & \myrr 0.3517   &   \myrr 0.2689    & \myrr 0.5984 \\
			\hline
		\end{tabular}
		\label{MTL_weighted_no_softmax}
		}
	\end{table}
\end{document}